\documentclass[a4paper,11pt,oneside]{article}

\usepackage[utf8]{inputenc}

\usepackage[left=2cm, right=2cm, bottom=2cm, top=2cm]{geometry}

\usepackage{mathtools}
\usepackage{amsmath, amsfonts, amssymb}

\usepackage{xcolor}

\usepackage{graphicx}

\usepackage{caption}
\usepackage{subcaption}

\usepackage{longtable}

\usepackage{hyperref}

\usepackage{enumerate}

\usepackage{tikz}

\usepackage{epstopdf}
\newcommand{\mypathsymne}{\rotatebox[origin=c]{45}{\large$\rightsquigarrow$}}

\newcommand{\mysetPoints}{P}
\newcommand{\mypoint}{p}
\newcommand{\mysubsetPointsTopo}{Q}
\newcommand{\mypointTopo}{q}
\newcommand{\mysetTopology}{T}
\newcommand{\mytopologicalSpace}{\mathcal{T}}

\newcommand{\mydirectedGraphsym}[1]{\mathcal{#1}}
\newcommand{\mysetVertices}[1]{V(#1)}
\newcommand{\mysetArcs}[1]{A(#1)}
\newcommand{\myfuncground}[1]{\operatorname{grnd}(#1)}
\newcommand{\myfuncnonground}[1]{\operatorname{nongrnd}(#1)}
\newcommand{\myfuncreciprocalground}[1]{\operatorname{grecip}(#1)}
\newcommand{\mymodifgraph}[1]{\tilde{#1}}

\newcommand{\mysetWalkVertices}[1]{\bar{V}(#1)}
\newcommand{\mysetWalkArcs}[1]{\bar{A}(#1)}
\newcommand{\mymechsymstemplate}[1]{\mathbb{#1}}
\newcommand{\mymechsym}{\mymechsymstemplate{M}}
\newcommand{\myconstituentsym}{\mymechsymstemplate{C}}
\newcommand{\myemergentsym}{\mymechsymstemplate{E}}

\newcommand{\mylistpropssym}{L}
\newcommand{\mylistpropsnode}{\mylistpropssym_\mathrm{node}}
\newcommand{\mylistpropsdilink}{\mylistpropssym_\mathrm{dilink}}

\newcommand{\myWalk}{W}
\newcommand{\myvertex}{v}

\title{A framework of defining, modeling, and analyzing cognition mechanisms}
\author{Amir Fayezioghani\thanks{Independent Researcher, Delft, The Netherlands\newline Email: amir.fayezi@gmail.com\newline \raisebox{-0.5em}{\includegraphics[height=1.5em]{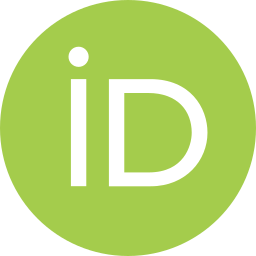}} \url{https://orcid.org/0000-0002-5694-6490}}}
\date{November, 2023}

\begin{document}
	
	\usetikzlibrary{graphs, shapes.geometric, shapes.arrows}

\maketitle

\begin{abstract}
	Cognition is a core part of and a common topic among philosophy of mind, psychology, neuroscience, AI, and cognitive science. Through a mechanistic lens, I propose a framework of defining, modeling, and analyzing cognition mechanisms. Firstly, appropriate terms are introduced and used in explanations related to the framework and within the definition of a mechanism. I implicitly contend that this terminology essentially characterizes a conceptual world required for discussions in this paper. Secondly, a mathematical model of a mechanism based on directed graphs is proposed.	Thirdly, the definition of a base necessary for a mechanism to be classified as a cognition mechanism is proposed. I argue that the cognition base has the features of the cognition self of humans. Fourthly, three ways to mechanistically look at mechanisms is defined and specific instances of them are suggested. Fifthly, standards for visualization and presentation of mechanisms, cognition mechanisms, and the instances to mechanistically look at them are suggested and used to analyze cognition mechanisms through appropriate examples. Finally, the features of this paper are discussed and prospects of further development of the proposed framework are briefly expressed.
	\vspace{1.5em}
	
\end{abstract}

\tableofcontents

\section{Introduction}
\emph{Cognition} is either \emph{computation}, \emph{representation}, or both by a \emph{mind}. I refer to mind as a system which has \emph{self} which is one of the key factors of characterizing the mind\footnote{For a review and better understanding of \emph{the mental}, refer to \cite{Pernu_2017}.}. A \emph{presentation} in a mind is a sub-state of the mind which is taken by the mind (internally, activated by an external signal, or from an extended cognition view) and is accessible to the mind's self. A representation in a mind is a presentation which is stored by the mind, where the \emph{storage} includes the mind's processes which maintains the presentation for a time longer than generation of the presentation and than accessing to the presentation by the mind's self\footnote{The definition of representation, here, requires definition of \emph{time} for the mind's self.}. Presentations and representations are \emph{teleological} or \emph{non-teleological} to the mind. Teleological presentations and representations are those which are used by the mind, where the \emph{usage} includes the mind's processes needed for its \emph{existential} and \emph{functional} purposes. Of course, if teleology is considered necessary for the definition of a mind, every state or act of the mind must be teleological, and thus, non-teleological presentations and representations will not be considered as the mind's actions. By adding more explanations to or by specifying terms of the definition of representation stated here, one can define more enriched or specific classes of representation in a mind.\footnote{For a recent classification of representations refer to \cite{Piccinini_2022b}.}

When a presentation taken by a mind is activated by the mind's self, I say the mind \emph{computes} (to the presentation) or a \emph{computation} is done by the mind. With this minimal definition, computation a) is independent of the storage of presentation by the mind, b) is functional (i.e.\ a function from something to a presentation in the mind), and c) is not trivial of presentation (i.e.\ does not include every presentation taken by the mind). Similar to representation, more enriched or specific computation could be defined and explained such as the computations defined and discussed by Chalmers~\cite{Chalmers_1995} and by Piccinini~\cite{Piccinini_2007}. It is worth noting that the domain of possible representations and possible computations in a mind are restricted to the boundaries necessitated by resources and capabilities of the mind. For example, a mind with finite time cannot compute to a result of the halting problem; or, a mind with $n$ possible distinct presentations cannot compute to every presentation of a mind with $m$ possible distinct presentations, where $m>n$.

I adopt a (new) mechanistic lens for definition of cognition. The New Mechanism or the new mechanical philosophy is a framework for thinking about the philosophical assumptions underlying many areas of science \cite{sep-science-mechanisms_2019}. The most commonly cited characterizations of the term ‘mechanism’ are proposed in \cite{machamer_darden_craver_2000}, \cite{glennan_2002}, and \cite{Bechtel_2005}. Each of the characterizations has four basic aspects, \emph{phenomena}, \emph{causings}, \emph{parts}, and \emph{organization} \cite{sep-science-mechanisms_2019}. I call `phenomenon and causing' \emph{emergents} of and `parts and organization' \emph{constituents} of a mechanism. A significant distinction between constituents and emergents is that constituents are substances while emergents are capabilities. Therefore, a substance is directly observable while a capability is indirectly observed via its \emph{samples}. It seems from the characterizations that, if possible at all, the aspects of a mechanism should be seen from a system's (either its or another system's) perspective. In this paper, I look at mechanisms from a human's perspective. Thus, I assume humans can observe, explain, and explicate a mechanism in order to understand its all aspects, emergents, and constituents, respectively. Glennan et al.\cite{Glennan_2022} have identified six theses from the results of philosophical investigations of as well as scientific search for the role of mechanisms.

Philosophy, psychology, neuroscience, Artificial Intelligence (AI), and cognitive science deal with cognition with different purposes, methods, and tools. Philosophy sees cognition as an action or product of the mind and investigates on it in order to better understand and explain the mind. Psychology observes behavior of people and collects people's expressions about their behavior so as to obtain knowledge based on information from the both sources of cognition. Neuroscience reduces internal and external actions (i.e.\ cognition) of living beings to their neural network's functions. Cognitive models are proposed in AI with the purpose of making tools which help better cognition or which extend cognition of us or other cognitive devices such as robots. And, cognitive science tries to integrate the mentioned knowledge produced from all these areas and disciplines. I look at cognition from bottom up while wearing a mechanistic lens, standing somewhere between neuroscience and psychology, and intending to clearly define a foundation of cognition; and, I attempted to maintain this mental configuration throughout this paper. In addition, cognition is seen by me as both the representation and the computation by a mind in general. Throughout the rest of this paper, unless specified otherwise, the term `cognition' is used instead of `semi-quasi-pseudo-cognition' which means that memory, (per)ception/(per)action, and will are not considered for the class of cognition systems under study. Hence, I define a \emph{cognition mechanism} as a mechanism for which it is possible a) to identify a base from its constituents and b) to verify a process which engages that base from its emergents. Moreover, a cognition mechanism is distinguished from a cognitive mechanism for which a base with will could be identified.

Proposing, suggesting, and using a proper terminology is crucial in conveying the massage of a text to potential readers. It is helpful to bear in mind that choosing a certain terminology must not change the conceptual world which is desired to be communicated but might give us a chance to rethink about the conceptual world and better understand or define it. Moreover, finding an existing terminology used in a domain of thought which (partially or totally) corresponds to another existing terminology used in another domain of thought is a strong evidence of existence of mutual concepts of the domains, which can lead to clearer communications between the domains. Another important key point in defining terms of a source terminology is that the process of definition by using a different or a mixed target terminology without giving an explicit relation between the source and the target is tricky since the result could be positive and cause better intuitive understandings for a group of readers while being negative and confuse another group of readers about desired concepts. Therefore, I propose different terminologies (which are mathematical, mechanistic, and cognitive) and give correspondence between them when defining terms used in the definition of mechanisms in Section~\ref{sec:math_defs}, Section~\ref{sec:mech_desc_model}, and Appendix~\ref{apdx:mech_terms_firstperson}. These terminologies similarly characterize the desired conceptual world though each of them may have a different a priori definition in the mind of a general reader.

Although different knowledge bodies are different in their logics and languages, they all are produced and cognizable by humans. Therefore, there is a good evidence that all humans have the same potential for cognition, which is also strengthened by neuroscientific findings. Moreover, in science, specially physical science, an entity in a level is composed of entities and their interactions in a lower level\footnote{Roughly speaking, if more than one entity compose an entity, it is said that the level of the former entities is lower than the level of the latter entity.}. As human knowledge grows and the technology based on it is developed, the physical science finds lower and lower levels of the physical world none of which can be claimed to be the lowest. On contrary, it is cognitively seen that the human cognition has a lowest level which is the level of (conscious) self. A compact description of this cognitive observation of human cognition is that each human has his/her own inseparable self which is single, unique, and basal, and which nothing is mentally more underlying, primitive, and unified than it. I propose a cognition mechanism with its cognition base at a lowest level which has the features of self in Section~\ref{sec:cognition_base}.

The ways a cognition mechanism may observe another cognition mechanism is defined and appropriate methods for analyses of mechanisms via such observations are proposed in Section~\ref{sec:metainfra_mechs}. In addition, core features of the framework of defining and analyzing cognition mechanisms are exemplified in Section~\ref{sec:examples}. These features are further discussed and prospects of the future development of the framework are expressed in Section~\ref{sec:discussions_prospects}.

\section{A framework of defining, modeling, and analyzing cognition mechanisms}
In this section, 1) some mathematical requisites are brought, 2) a definition of mechanisms from an epistemic third-person view is provided by using an appropriate terminology, and a model of mechanisms is given, 3) a definition of a base suitable for cognition is proposed, and the model of cognition mechanisms are given, and 4) meta-/infra-/iso-mechanisms are introduced and methods of characterization, substancization, and formization of mechanisms are proposed.

\subsection{Some required assumptions and definitions of mathematical terms and objects}   \label{sec:math_defs}
Let $\mysetPoints$ be a \emph{set}\footnote{Any mathematical set referred to in this paper is a mathematical set without repetition of its members.} of points. A \emph{neighborhood topology} $\mysetTopology$ on $\mysetPoints$ is defined as a set of pairs of $(\mypoint, \mysubsetPointsTopo)$, where $\mypoint \in \mysetPoints$ and $\mysubsetPointsTopo \subseteq \mysetPoints$. A \emph{neighborhood topological space} $\mytopologicalSpace$ is defined as the set of points $\mysetPoints$ equipped with the set of pair $\mysetTopology$. Similarly, a \emph{neighborhood topological spatium} (usually called a neighborhood topological point) is defined as a point $\mypoint$ equipped with a pair of $(\mypoint, \mysubsetPointsTopo)$.

Assume that a) all sub-sets of $\mysetPoints$ are countable and b) any pair of $(\mypoint, \mysubsetPointsTopo)$ can be substituted for a set of pairs of $(\mypoint, \mypointTopo)$ for all $\mypointTopo \in \mysubsetPointsTopo$. With these assumptions, the neighborhood topological space becomes a \emph{directed graph} (or \emph{digraph} in short). I adhere to the stated terminology as well as to a common terminology used in (Directed) Graph Theory\footnote{Most of the terms and definitions of directed graphs in this section are in reference to \cite{Bang-Jensen_2009}.} in the modeling of mechanisms.

A digraph $\mydirectedGraphsym{D}$ is a set of points called the \emph{set of vertices} $\mysetVertices{\mydirectedGraphsym{D}}$ associated with the set of pairs of those points called the \emph{set of arcs} $\mysetArcs{\mydirectedGraphsym{D}}$. A digraph $\mydirectedGraphsym{H}$ is a \emph{sub-digraph} of a digraph $\mydirectedGraphsym{D}$ if $\mysetVertices{\mydirectedGraphsym{H}} \subseteq \mysetVertices{\mydirectedGraphsym{D}}$, $\mysetArcs{\mydirectedGraphsym{H}} \subseteq \mysetArcs{\mydirectedGraphsym{D}}$, and every arc in $\mysetArcs{\mydirectedGraphsym{H}}$ has both end-vertices (i.e.\ both elements of the pair indicating each arc) in $\mysetVertices{\mydirectedGraphsym{H}}$. Digraph $\mydirectedGraphsym{D}$ is the \emph{union} of digraphs $\mydirectedGraphsym{H}$ and $\mydirectedGraphsym{L}$ if $\mysetVertices{\mydirectedGraphsym{D}} = \mysetVertices{\mydirectedGraphsym{H}} \cup \mysetVertices{\mydirectedGraphsym{L}}$ and $\mysetArcs{\mydirectedGraphsym{D}} = \mysetArcs{\mydirectedGraphsym{H}} \cup \mysetArcs{\mydirectedGraphsym{L}}$. A \emph{walk} in $\mydirectedGraphsym{D}$ is a sequence of vertices $\myWalk = \myvertex_1 \, \myvertex_2 \, \myvertex_3 \, \ldots \, \myvertex_k$ in which all pairs of consecutive vertices $(\myvertex_i, \myvertex_{i+1})$ for $1 \leq i \leq k-1$ is in $\mysetArcs{\mydirectedGraphsym{D}}$. The set of vertices and the set of arcs which are \emph{traversed} via a walk $\myWalk$ are shown by $\mysetWalkVertices{\myWalk}$ and $\mysetWalkArcs{\myWalk}$, respectively. It is worth highlighting that returning to some of the previously traversed vertices and arcs during a walk $\myWalk$ does not add extra members to the sets $\mysetWalkVertices{\myWalk}$ and $\mysetWalkArcs{\myWalk}$. A digraph composed of these sets is called the \emph{traversed sub-digraph} (of that walk). In $\myWalk$, $\myvertex_1$ and $\myvertex_k$ are called \emph{initial} and \emph{terminal} vertices, respectively, and the rest of vertices are called \emph{medial}; and, both $\myvertex_1$ and $\myvertex_k$ are \emph{end-vertices} (of $\myWalk$). If there can be a walk from an initial vertex $x$ to a terminal vertex $y$, it is said that $x$ is \emph{connected to} $y$ or $y$ is \emph{connected from} $x$\footnote{In some texts, the term `approachable' is used instead of `connected'.}. $\myWalk$ is \emph{closed} if $\myvertex_1 = \myvertex_k$; otherwise, it is \emph{open}. A \emph{path} (from $\myvertex_1$ to $\myvertex_k$) is a walk $\myWalk$ ($= \myvertex_1 \, \myvertex_2 \, \ldots \, \myvertex_k$) whose all vertices are distinct. If $\myvertex_1$, $\myvertex_2$, \ldots, $\myvertex_{k-1}$ are distinct, $k \geq 3$, and $\myvertex_1 = \myvertex_k$ in a walk, the walk is called a \emph{cycle} (over $\myvertex_1$ or $\myvertex_k$). If there is an arc $(\myvertex_i, \myvertex_i)$ in a digraph, it is called a \emph{loop} (of $\myvertex_i$). Accordingly, a \emph{simple loop} (of $\myvertex_i$) can be shown by the walk $\myWalk = \myvertex_i \, \myvertex_i$. A walk $U = x_1 \, x_2 \, \ldots \, x_n$ can be succeeded by a walk $V = y_1 \, y_2 \, \ldots \, y_m$ in order to make a walk $\myWalk = x_1 \, x_2 \, \ldots \, x_{n-1} \, z \, y_2 \, \ldots \, y_m$ only if $x_n = y_1 = z$; this is called \emph{succession} of walks. An \emph{all-path} is the set of all possible paths from a vertex to another vertex of a digraph and, similarly, an \emph{all-cycle} is the set of all possible cycles over a vertex of a digraph. The traversed sub-digraph of an all-path (or an all-cycle) is assumed to be equal to the union of traversed sub-digraphs of all paths (or all cycles) of that all-path (or that all-cycle) and is called an \emph{all-pathic sub-digraph} (or \emph{all-cyclic sub-digraph}). In a digraph $\mydirectedGraphsym{D}$ and for a vertex $x$ of it, if there can be at least a cycle over $x$, $\mydirectedGraphsym{D}$ is \emph{cyclic} over $x$. If $\mydirectedGraphsym{D}$ is cyclic over all its vertices, it is \emph{strongly cyclic}. In a digraph $\mydirectedGraphsym{D}$ and for every two arbitrary vertices $x$ and $y$ of it, if $x$ is connected to $y$ and $x$ is connected from $y$, $\mydirectedGraphsym{D}$ is \emph{strongly connected}. A strongly connected digraph is strongly cyclic but a strongly cyclic digraph is not necessarily  strongly connected.

\subsection{Definition and mathematical model of a mechanism}   \label{sec:mech_desc_model}
In the definition of a mechanism here, an appropriate terminology for cognition mechanisms from an epistemic third-person view is suggested and used. In order to mathematically model a mechanism, it suffices to give a clear unambiguous correspondence between the used terms and their relations and mathematical objects and their relations. However, some meta-lingual terms which may be modeled by meta-mathematical objects are used for the sake of ease of reading and understanding.

As mentioned in the introduction, constituents of a mechanism are parts and the organization of parts. `Part' and `organization' are general terms which are replaced by `spot' and `proximities' in the definitions of a mechanism, respectively. Thus, constituents of a mechanism are \emph{spots} and their \emph{proximities}. For each spot, there is a \emph{vicinity} which is all spots being proximate to it. A spot with its vicinity is called an \emph{ensemble} and a group of ensembles is called an \emph{assembly}. If spots and their proximities are \emph{countable}, they are respectively called a group of \emph{nodes} and a group of \emph{direct links}. I postulate that a) all spots and their proximities are countable and b) there can always be a direct link from a node to each of its vicinity's nodes. With these \emph{constitutive postulates}, an assembly definitionally becomes a \emph{network}. A \emph{net} is defined as a portion of a network\footnote{Notice that a portion of a network is either nothing, a piece of the network, or the entire network.}. A \emph{unification} of nets is defined as a net composed of a group of all the nets' nodes together with a group of all the nets' direct links. If a mechanism does things and if we, as observers of the mechanism, can observe these things, we can identify a phenomenon of the mechanism based on a certain collection of the things. Here, a mechanism is assumed without any interactions with other mechanisms and consequently without causings. Thus, emergents of a mechanism are restricted to only phenomena. From a meta-cognitive view, I assume that it is intrinsically possible for the mechanism to pass something or to let something pass between its nodes via their direct links to each other and that we can see and sequentially follow this process. In this situation, it is said that the mechanism \emph{intramits} or performs an \emph{intramission}. Similar to part and organization, `phenomenon' is a general term which is implicitly interchanged with the `capacity of a mechanism to intramit', and `sample of the phenomenon' is replaced by `intramission' in the remaining definitions of a mechanism. An intramission is defined as a sequence of nodes passing something to each other via direct links. In an intramission, the node from which the process starts and the node to which the process ends are called the \emph{initial} and the \emph{terminal} nodes, respectively; all other nodes are \emph{medial}. If it is possible for the mechanism to perform an intramission from a certain initial node to a certain terminal node, it is said that the first node is \emph{linked to} the second node or the second node is \emph{linked} from the first node. If the initial and the terminal nodes of an intramission are the same, that intramission is called a \emph{circulation} (over the initial or the terminal node); otherwise, it is a \emph{deliveration} (from the initial to the terminal node). If the nodes of a circulation except the terminal node or the nodes of a deliveration are distinct from each other, that circulation or deliveration is \emph{simple}. If one identifies the net in which an intramission is performed, that net is called the \emph{carrying net} (of the intramission). An intramission can be joined to another intramission in order to make a new intramission only if the terminal node of the first intramission is the same as the initial node of the second intramission. The procedure of \emph{joining} of intramissions is done by performing the first intramission and, immediately after its termination, performing the second intramission. In a network, a) a \emph{unit} is a net which is the unification of carrying nets of all possible simple circulations over a certain node of the network and b) a \emph{uniter} is a net which is the unification of carrying nets of all possible simple deliverations from a certain node to another certain node of the network. A network is \emph{(simple-)circulational over a node} if at least a (simple) circulation over that node can be performed in a net of it; and, a network is \emph{(simple-)deliverational from a node to another node} if at least a (simple) deliveration from that node to the other node can be performed in a net of it. If a network is (simple-)circulational over all its nodes, it is \emph{strongly (simple-)circulational}. In a network and for every two arbitrary nodes of it, if the nodes are linked to each other, that network is \emph{strongly linked}. A strongly linked network is strongly simple-circulational but a strongly simple-circulational network is not necessarily strongly linked. 

The used terminology is correspondent with the mathematical terminology as in Table~\ref{tab:cog_terminology_thirdperson}. Therefore, the model of a mechanism is constructed by substituting the mathematical terms in Section~\ref{sec:math_defs} for the used terms of mechanisms in this section. The above definitions of a mechanism are brought to give a better understanding of mechanisms; however, its mathematical model is more rigorous and should be referred to in case of any confusions.
\begin{table}
	\centering
	\caption{The used epistemic third-person terminology in the definition of a mechanism corresponding to the mathematical terminology}   \label{tab:cog_terminology_thirdperson}
	\begin{tabular}{p{6cm} | p{6cm}}
		\hline
		Mechanistic term & Mathematical term\\
		\hline
		spot & point\\
		proximity & (mathematical) relation in the form of pairs of objects\\		
		vicinity & neighborhood topology pair\\
		group of vicinities & neighborhood topology\\
		ensemble & neighborhood topological spatium\\
		assembly & neighborhood topological space\\
		node & vertex\\
		direct link & arc\\
		network & digraph\\
		net & sub-digraph\\
		intramission & walk\\
		circulation & closed walk\\
		deliveration & open walk\\
		simple circulation & cycle\\
		simple deliveration & path\\
		carrying net of an intramission & traversed sub-digraph of a walk\\
		unification & union\\
		unit & all-cyclic sub-digraph\\
		uniter & all-pathic sub-digraph\\
		(strongly) simple-circulational & (strongly) cyclic\\
		(strongly) linked & (strongly) connected\\
	\end{tabular}
\end{table}
For interested readers, the definition of a mechanism with a terminology from an epistemic first-person view and a model for mechanisms is also suggested in Appendix~\ref{apdx:mech_terms_firstperson}.

\subsubsection{Some clarifying characteristics of mechanisms}
According to Section~\ref{sec:mech_desc_model}, constituents of a mechanism are spots and proximities which are respectively modeled by points and mathematical relations; or, when the constitutive postulates are considered for a mechanism, constituents of a mechanism are nodes and direct links which are respectively modeled by vertices and arcs. And, emergents of a mechanism (including only phenomena) are intramissions which are modeled by walks in digraphs.

Now, assume that someone, like me in this paper, adopts a mechanistic lens and looks at intramissions of a mechanism. Then, he/she defines the intramissions as a mechanism by starting from seeing its constituents. Therefore, an intramission will constitutively be a network in his/her eyes, which I called it a carrying net of the intramission. Also, the intramission will emergively (including only phenomenally) be an intramission of this network. I call this process \emph{mechanation} which is composed of \emph{conception of instances} and \emph{instantiation of concepts}\footnote{Refer to the definition of mechanation, conception, and instantiation in Appendix~\ref{apdx:mech_terms_firstperson} for a better understanding of them.}. To more specify the process of mechanation and the construction of mechanisms, some characteristics of the process are introduced and meta-cognitively explained and justified in the following:
\begin{enumerate}
	\item   \emph{Discrete} Conception: It is intuitively easier to conceive instances discretely.   \label{charsMechantn:discrete}
	
	\item   \emph{Restricted} Conception: Cognition and accordingly conception naturally use limited resources and energy.   \label{charsMechantn:restricted}
	
	\item   \emph{Uncut} Conception: Conception naturally includes dealing with spots and their proximities together and not separately, which is related to (active/passive) generation of information by a cognition mechanism.   \label{charsMechantn:uncut}
	
	\item   \emph{Non-trivial} Conception: Conception is not mere existence of spots and proximities; it is about the relation among distinct spots by their proximities.   \label{charsMechantn:non-trivial}
	
	\item   \emph{Uni-memorial} Conception: Cognition and accordingly conception by a mechanism are defined with the aid of one integrated memory of us humans.   \label{charsMechantn:uni-memorial}
	
	\item   \emph{Thorough} Conception: Conception is about the group of all vicinities and not a sub-group of such.   \label{charsMechantn:thorough}
	
	\item   \emph{Comprehensive} Conception: Conception is about the group of all spots together with their vicinities and not a sub-group of such.   \label{charsMechantn:comprehensive}
	
	\item   \emph{Maximal} Conception: Conception naturally tends to maximize information per memory by giving access from each node to every nodes in its vicinity.   \label{charsMechantn:maximalistic}
	
	\item   \emph{Classic} Conception: Commonsense of humans is that conception is about nodes and direct links each of which relate two nodes.   \label{charsMechantn:classic}
	
	\item   \emph{Single-leveled} Conception: Intuitively, conception is performed in a single step. In addition, with a mechanistic lens, humans (who conceive) are seen as mechanisms. Therefore, these two premises imply that conception should be single-leveled\footnote{Assume that a first human conceives a mechanism in a step in a level (of fundamentality). Thus, a second human should conceive the first human in two steps (i.e.\ one step to conceive the first human in a level and one step to conceive the first human's conception of the mechanism in a lower level). This process may continue for conceptions of a third, a fourth, \ldots human of his/her previous human and his/her conception of his/her previous human and so forth. Therefore, the conception will be performed in multi-steps. In other words, a multi-level conception implies performing a multi-step conception, which is the contrapositive proof of the statement. Moreover, it can also be concluded from a multi-step conception that a conception will probably be practically incomplete in the sense that it should be cut after a limited number of steps due to the limited resources of humans versus the number of all of the entities and humans in the history of life.}.   \label{charsMechantn:single-leveled}
	
	\item   \emph{Actual} Instantiation: Instantiation is practically an action and thus it is actual.   \label{charsMechantn:actual}
	
	\item   \emph{Diversal} Mechanation: Mechanation naturally tends to maximize information per try by avoiding similar nets.   \label{charsMechantn:diversal}
	
	\item   \emph{Steady} Mechanation: Mechanation of emergents of a mechanism can recursively be performed til infinity. In other words, conception of instances and instantiation of concepts can consecutively be performed for an infinite number of times. However, there is a technique based on imposing self-reference to an intramission that assures us that nothing iteratively non-trivial will be found in the mentioned process of mechanistic conception. The technique is to assume that an intramission is equal to the only intramission of its carrying net (as a network). Therefore, the intramission can solely be either a simple circulation or a simple deliveration which is irreducible to any other type of intramission.   \label{charsMechantn:steady}
	
	\item   \emph{Exhaustive} Mechanation: In a conception process, there may be a node whose vicinity includes more than one node distinct from that node, and consequently an instantiation may construct an intramission which takes either of the direct links from that node to its vicinity's nodes. An exhaustive mechanation considers all possible choices in its instantiation.   \label{charsMechantn:exhaustive}
\end{enumerate}

A summary of the characteristics of a mechanism which the explained meta-cognitive characteristics of a mechanation amount to are listed in Table~\ref{tab:cog_constraints} in mechanistic and mathematical terms.
\begin{table}
	\centering
	\caption{The characteristics of a mechanism corresponding to characteristics of a mechanation in mechanistic and mathematical terms}   \label{tab:cog_constraints}
	\small
	\begin{tabular}{| l | p{7cm} | p{8cm} |}
		\hline
		\# & In mechanistic terms, in a mechanism, \ldots & In mathematical terms, in a topological space or in a digraph, \ldots\\
		\hline
		\ref{charsMechantn:discrete} & spots and proximities are \emph{individual} & each point and each mathematical relation are single mathematical objects\\
		\hline
		\ref{charsMechantn:restricted} & spots and proximities are \emph{confined} & the set of points and the set of mathematical relations are bounded\\
		\hline
		\ref{charsMechantn:uncut} & every vicinity is \emph{existent} & the second element of every pair of neighborhood topology is a non-empty sub-set of points\\
		\hline
		\ref{charsMechantn:non-trivial} & every spot is \emph{non-isolated} & every point of the set of points is either a) the first element of at least a pair of neighborhood topology whose second element has at least a point that is distinct from that first element or b) a member of the second element of at least a pair of neighborhood topology whose first element is distinct from that member.\\
		\hline
		\ref{charsMechantn:uni-memorial} & the group of vicinities is \emph{non-multial} & if first elements of every two pairs of neighborhood topology are the same, the second elements of those pairs must also be the same\\
		\hline
		\ref{charsMechantn:thorough} & the group of vicinities is \emph{covering} & the set of first elements extracted from the set of pairs of neighborhood topology is equal to the set of points\\
		\hline		
		\ref{charsMechantn:comprehensive} & the assembly is \emph{covering} & the underlying set of points of neighborhood topological space is equal to the set of points\\
		\hline
		\ref{charsMechantn:maximalistic} & every direct link is \emph{uniformly dispersive} & all pairs of neighborhood topology are distributive relations from the first element of each pair to every points of the second element of that pair\\
		\hline
		\ref{charsMechantn:classic} & every direct link is \emph{two-ended} with \emph{single} ends & every arc is a binary relation from a vertex to a vertex\\
		\hline
		\ref{charsMechantn:single-leveled} & every network is \emph{universal} & every digraph is solely composed of vertices and arcs each of which has a certain fixed definition\\
		\hline
		\ref{charsMechantn:actual} & every intramission is \emph{accordant to a net} of the network & every walk in a digraph is definitionally a sequence of vertices which is composed of consecutive vertices according to the set of arcs\\
		\hline
		\ref{charsMechantn:diversal} & intramissions are \emph{non-extra} and\newline carrying nets are \emph{distinct} & any sub-digraph of the digraph is the traversed sub-digraph of no more than one walk\\
		\hline
		\ref{charsMechantn:steady} & intramissions are \emph{simple} and\newline carrying nets are \emph{plain} & any walk is either a path or a cycle\\
		\hline
		\ref{charsMechantn:exhaustive} & intramissions are \emph{sweeping} and\newline carrying nets are \emph{plural} & there should be all walks between every two (distinct or same) vertices\\
		\hline
	\end{tabular}
\end{table}

To clarify and disambiguate the term `mechanism', it is worth highlighting that characteristics \#\ref{charsMechantn:discrete} to \#\ref{charsMechantn:actual} are explicitly or implicitly considered in the definitions of as well as the mathematical definitions related to the model of a mechanism; characteristics \#\ref{charsMechantn:discrete} and \#\ref{charsMechantn:restricted} imply that spots and proximities are \emph{countable}\footnote{Countable spots and countable proximities are definitionally called a group of nodes and a group of direct links, respectively}; the countability of spots and proximities as well as characteristic \#\ref{charsMechantn:maximalistic} are stated as the constitutive postulates; characteristics \#\ref{charsMechantn:diversal}, \#\ref{charsMechantn:steady}, and \#\ref{charsMechantn:exhaustive} will be used in mechanistic characterizations and formizations of mechanisms defined in Section~\ref{sec:metainfra_mechs:mech_formiz} and exemplified in Section~\ref{sec:examples}.

\subsection{A base of a mechanism suitable for cognition}   \label{sec:cognition_base}
As mentioned in the introduction, existence of a base and the process(es) related to it is necessary in a mechanism to be classified as a cognition mechanism. This base is called \emph{cognition mechanism's base} (or \emph{cognition base} in short). I introduce a cognition base with a main postulate: A cognition base is a cognition mechanism. This postulate means that a cognition base is \emph{embodied} (similar to a cognition mechanism). If a cognition mechanism's network is universal and it has an embodied cognition base, the cognition base must be a sub-mechanism of it, and thus the cognition base is also \emph{embedded}. Therefore, a mechanism will be divided into a base and a non-base. In addition, I propose characteristics of an embedded embodied cognition base in mechanistic terms as well as characteristics of a sub-digraph as its model in mathematical terms in Table~\ref{tab:compcogbase_constraints}.
\begin{table}
	\centering
	\caption{Proposed characteristics of an embedded embodied cognition base in mechanistic and a ground in mathematical terms}   \label{tab:compcogbase_constraints}
	\newcounter{BaseCharCount}
	\small
	\begin{tabular}{| l | p{7.5cm} | p{7.5cm} |}
		\hline
		\# & In mechanistic terms, in a cognition mechanism, \ldots & In mathematical terms, in a digraph, \ldots\\
		\hline
		\refstepcounter{BaseCharCount} \label{char:cog_base:underlying} \theBaseCharCount & the base's network is \emph{fully linked from} the non-base's network & \textbf{every} vertices of the sub-digraph is connected from \textbf{all} vertices not in the sub-digraph\\
		\hline
		\refstepcounter{BaseCharCount} \label{char:cog_base:primitive} \theBaseCharCount & the base's network is \emph{not at all linked to} the non-base's network & \textbf{every} vertices of the sub-digraph is connected to \textbf{no} vertex not in the sub-digraph\\
		\hline
		\refstepcounter{BaseCharCount} \label{char:cog_base:nontriv_egalitarian_linked} \theBaseCharCount & the base's network is \emph{non-trivially egalitarianly linked to and from} a net & \textbf{every} vertices of the sub-digraph is connected to and from \textbf{all} vertices of a \textbf{certain} \textbf{non-enmpty} set of vertices in the digraph\\
		\hline
	\end{tabular}
\end{table}
These characteristics imply that the embedded embodied cognition base must be \emph{self-based}; that is, if an embedded embodied cognition base is considered as a mechanism with a universal network, its embedded embodied cognition base is itself. A self-based embedded embodied cognition base is called a \emph{cognition self} (or \emph{self} in short or \emph{ground} in mathematical terms) and the rest of the cognition mechanism is called \emph{cognition non-self} (or \emph{non-self} in short or \emph{non-ground} in mathematical terms).\footnote{It is worth reminding that `cognition', here, refers to `semi-quasi-pseudo-cognition' and, accordingly, `self' refers to `semi-quasi-pseudo-self'.} Cognition self's and cognition non-self's network have a group of nodes in common which are called the group of \emph{self-mutual} nodes of a cognition mechanism in mechanistic terms and the set of \emph{ground-reciprocal} vertices of a digraph in mathematical terms. Moreover, a cognition self's network is strongly linked and thus, strongly simple-circulational.

Now, I relate the features of human self brought in the introduction to some of the constitutive properties of a cognition self in Table~\ref{tab:humancog_vs_cogmech:constitutive} in case of modeling of the human cognition by a cognition mechanism.
\begin{table}
	\centering
	\caption{The list of features of human self which are satisfied by constitutive properties of a cognition self when modeling the human cognition by a cognition mechanism}   \label{tab:humancog_vs_cogmech:constitutive}
	\small
	\begin{tabular}{| l | l |}
		\hline
		\textbf{Feature of human self} & \textbf{Constitutive property of cognition self}\\
		\hline
		ownedness & embodiment\\
		\hline
		inseparability & embeddedness\\
		\hline
		underlyingness & characteristic~\#\ref{char:cog_base:underlying}\\
		\hline
		primitiveness & characteristic~\#\ref{char:cog_base:primitive}\\
		\hline
		unifiedness & characteristic~\#\ref{char:cog_base:nontriv_egalitarian_linked}\\
		\hline
		basality & characteristics~\#\ref{char:cog_base:underlying} and \#\ref{char:cog_base:primitive} together\\
		\hline
		singleness & characteristics~\#\ref{char:cog_base:underlying} and \#\ref{char:cog_base:nontriv_egalitarian_linked} together\\
		\hline
		uniqueness & characteristics~\#\ref{char:cog_base:primitive} and \#\ref{char:cog_base:nontriv_egalitarian_linked} together\\
		\hline
		compositional fundamentality & characteristics~\#\ref{char:cog_base:underlying}, \#\ref{char:cog_base:primitive}, and \#\ref{char:cog_base:nontriv_egalitarian_linked} together\\
		\hline
	\end{tabular}
\end{table}
In addition, again in modeling the human cognition by a cognition mechanism, there are features of human cognition which are satisfied by the emergive properties of a cognition mechanism depicted in Table~\ref{tab:humancog_vs_cogmech:emergive}.
\begin{table}
	\centering
	\caption{The list of features of human self which are satisfied by emergive properties of a cognition self when modeling the human cognition by a cognition mechanism}   \label{tab:humancog_vs_cogmech:emergive}
	\small
	\begin{tabular}{| p{7.5cm} | p{7.5cm} |}
		\hline
		\textbf{Feature of human self} & \textbf{Emergive property of cognition self}\\
		\hline
		Change of a human's perspective and/or speculation by himself/herself without learning new things does not change his/her cognition base. & An intramission does not change a network.\\
		\hline
		A human can change his/her perspective and/or speculation for an infinite number of times\footnote{This statement does not mean that all of these infinite perspectives are distinct. In contrast, the number of distinct perspectives is finite for a human because of limited mental resources of humans.} & There can be an infinite number of intramissions.\\
		\hline
		A human, in general, cannot speculate anything except himself/herself. & Any intramission initiated from a node of cognition self may only terminate to a node of cognition self.\\
		\hline
		A human can potentially speculate his/her whole self. & There can be an intramission between any two nodes of cognition self.\\
		\hline
	\end{tabular}
\end{table}

\subsection{Meta-/ifra-/iso-mechanisms}   \label{sec:metainfra_mechs}
Assume a mechanism $\mymechsym_i$ composed of constituents $\myconstituentsym_i$ and emergents $\myemergentsym_i$ in level $i$ each of which, in general, have \emph{properties}\footnote{`Properties', here, refers to a mixed collection of first-order as well as higher-order properties.}. If one transforms $\mymechsym_i$ to constituents $\myconstituentsym_{i+1}$ and employs emergents $\myemergentsym_{i+1}$, he/she has performed a mechanation from $\mymechsym_i$ to $\mymechsym_{i+1}$, a meta-mechanation from level $i$ to $i+1$, a meta-mechanation on level $i$, or just a meta-mechanation when level $i$ is implicitly stated in a context. A meta-mechanation from level $i$ to $i+1$ leads to the \emph{meta-mechanism} $\mymechsym_{i+1}$ and is such that, in general, $\myconstituentsym_{i+1}$ will be different from $\myconstituentsym_i$ and with extended properties. In this case, the portion of the meta-mechanation related to $\myemergentsym_i$ is responsible for the `extended'. On the other hand, if one transforms $\myconstituentsym_i$ to $\mymechsym_{i-1}$ and dismisses $\myemergentsym_{i}$, he/she has performed a mechanation from $\mymechsym_i$ to $\mymechsym_{i-1}$, an infra-mechanation from level $i$ to $i-1$, an infra-mechanation on level $i$, or just an infra-mechanation when level $i$ is implicitly stated in a context. An infra-mechanation from level $i$ to $i-1$ leads to the \emph{infra-mechanism} $\mymechsym_{i-1}$ and is such that, in general, $\myconstituentsym_{i-1}$ will be different from $\myconstituentsym_i$ and with reduced properties. In this case, the portion of the infra-mechanation related to $\myemergentsym_{i-1}$ is responsible for the `reduced'. And, as a last case, if one transforms $\mymechsym_a$ to $\mymechsym_b$ (both of which are in the same level, say level $i$), he/she has performed a mechanation from $\mymechsym_a$ to $\mymechsym_b$, an iso-mechanation from form $a$ to $b$, or just an iso-mechanation when form $a$ and $b$ are implicitly stated in a context. An iso-mechanation from form $a$ to $b$ leads to the \emph{iso-mechanism} $\mymechsym_b$ and is such that, in general, $\myconstituentsym_b$ will be similar to $\myconstituentsym_a$ but with more/less properties while $\myemergentsym_b$ will be similar $\myemergentsym_a$ but with less/more properties. In this case, the portions of the iso-mechanation related to $\myemergentsym_a$ and $\myemergentsym_b$ are responsible for the `more'/`less' and the `less'/`more'. Hence, an iso-mechanation from form $a$ to $b$ in level $i$ is equivalent to a meta-mechanation on level $i$ followed by an infra-mechanation on level $i+1$. As a convention, in an iso-mechanation from form $a$ to $b$, if $\myconstituentsym_b$ has more/less properties than $\myconstituentsym_a$, the iso-mechanation and consequently the iso-mechanism is \emph{higher-order}/\emph{lower-order}.

For example, a \emph{mechanistic characterization}\footnote{characterization as complete identification} of a mechanism may be done by a meta-mechanation leading to all of its units and uniters and by employing appropriate emergents on them;\footnote{Some terms related to using units/uniters in mechanistic characterizations in different contexts are mechanical/ dynamic system theoretic, symbolic (symbolistic)/ connectionic (connectionistic), definitive/ descriptive, and ceptive/ active.} or, a \emph{mechanistic substancization}\footnote{substancization as complete realization} of (constituents of) a mechanism may be done by an infra-mechanation leading to its network's nodes without labels and with some properties and the process of labeling them as an emergent on the nodes; or, a \emph{mechanistic formization}\footnote{formization as equivalent formation} of a mechanism may be done by an iso-mechanation leading to all its nodes completely linked to each other each of which without label and with some properties and the the process of labeling them based on the properties as an emergent.

\subsubsection{Mechanistic characterization, substancization, and formization of mechanisms}   \label{sec:metainfra_mechs:mech_formiz}
Each of mechanistic characterization, substancization, and formization of mechanisms can be performed in different ways. Firstly, I propose a mechanistic characterization of a mechanism $\mymechsym_{i}$ by a meta-mechanation which leads to a meta-mechanism $\mymechsym_{i+1}$ whose
constituents are nodes of $\mymechsym_{i}$ equipped with units and/or uniters of $\mymechsym_{i}$. A mechanistic characterization \emph{mode} based on only units, only uniters, and both units and uniters are usually called a \emph{symbolistic}, a \emph{connectionistic}, and a \emph{hybridistic} characterization, respectively. It is experientially seen that symbolic (connectionic) characterizations are better understood ceptionally (actionally). Moreover, all mechanisms can be connectionistically characterized while not all mechanisms can be symbolistically characterized. Roughly speaking, the reason is that uniters bear all information of direct and indirect links between nodes while units cannot necessarily do so. Therefore, an strategy of hybridistic characterization could be identifying all possible units and then identifying a sufficient number of uniters that complete the characterization. It is worth reminding that a mechanism with a strongly linked network can be characterized either symbolistically, connectionistically, or hybridistically. Secondly, I propose a mechanistic substancization of a mechanism $\mymechsym_{i}$ by an infra-mechanation which leads to an infra-mechanism $\mymechsym_{i-1}$ whose constituents are unlabeled nodes of $\mymechsym_{i}$ and whose emergent is a process of assigning labels to the nodes. Thirdly, I propose a higher-order mechanistic formization of a mechanism $\mymechsym_a$ by an iso-mechanation which leads to an iso-mechanism $\mymechsym_b$ whose
\begin{itemize}
	\item   $\myconstituentsym_b$ is a network constructed by
	\begin{itemize}
		\item   the same nodes as $\mymechsym_a$'s network but without labels
		\item   directly linking its all nodes to its all nodes
	\end{itemize}
	such that each node and each direct link has a list of property measures $\mylistpropsnode$ and $\mylistpropsdilink$ of Table~\ref{tab:props_nodesdirectlinks} on the corresponding nodes of $\mymechsym_a$'s network according to one of the following \emph{modes}:
	\begin{itemize}
		\item[]   \emph{single-leveled}: $\mylistpropsnode$ is the property measure \#\ref{numcycles}, and $\mylistpropsdilink$ is the property measure \#\ref{numpaths}.   \label{mech_formiz_uniter_unit}
		
		\item[]   \emph{mixed-leveled}: $\mylistpropsnode$ is a list of the property measures \#\ref{numneighbornodes:out} and \#\ref{numneighbornodes:in}, and $\mylistpropsdilink$ is the property measure \#\ref{numpaths}. \label{mech_formiz_node_uniter}
	\end{itemize}

	\item   $\myemergentsym_b$ is the process of assigning labels to the network's nodes and direct links based on their property measures.
\end{itemize}
\begin{table}
	\centering
	\caption{Some measures of nodes and direct links of a mechanism in mathematical terms.}   \label{tab:props_nodesdirectlinks}
	\newcounter{NodeLinkPropCount}
	\begin{tabular}{| l | l |}
		\hline
		\textbf{\#} & \textbf{Property Measure}\\
		\hline
		\refstepcounter{NodeLinkPropCount} \label{numneighbornodes:out} \theNodeLinkPropCount & the number of all vertices adjacent from vertex $x$ in a digraph\\
		\hline
		\refstepcounter{NodeLinkPropCount} \label{numneighbornodes:in} \theNodeLinkPropCount & the number of all vertices adjacent to vertex $x$ in a digraph\\
		\hline
		\refstepcounter{NodeLinkPropCount} \label{numcycles} \theNodeLinkPropCount & the number of all cycles over vertex $x$ in a digraph\\
		\hline
		\refstepcounter{NodeLinkPropCount} \label{numpaths} \theNodeLinkPropCount & the number of all paths from vertex $x$ to vertex $y$ in a digraph\\
		\hline
	\end{tabular}
\end{table}
It is observed from the proposed formization that a) it uses units and uniters of $\mymechsym_a$ when determining property measures \#\ref{numcycles} and \#\ref{numpaths} for its nodes and direct links, respectively and b) it determines a process of assigning labels to its unlabeled nodes. These observations implicitly imply that the proposed higher-order mechanistic formization is equivalent to the proposed mechanistic characterization followed by the proposed mechanistic substancization.

\section{Examples of a cognition mechanism and a cognition base}   \label{sec:examples}
This section exemplifies mechanisms and cognition mechanisms with proper visualizations of their models as well as analyzes their properties and gives deeper insight into their implications.

\subsection{Visualization of mechanisms and cognition mechanisms}   \label{sec:example1}
This example provides the reader with a visualization type of mathematical models of constituents of mechanisms in general as well as cognition mechanisms in particular.

\begin{figure}
	\centering
	
	\begin{subfigure}[b]{0.45\textwidth}
		\centering
		
		\input{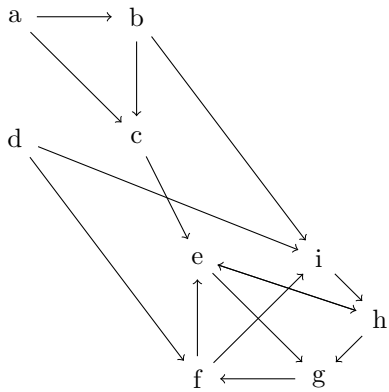}
		
		\caption{A digraph is colored in black in general. Vertices of a digraph are indicated by their labels and arcs of a digraph are drawn by arrows between the vertices.}   \label{fig:example1:digraph}
	\end{subfigure}
	\hspace{0.0cm}
	\begin{subfigure}[b]{0.45\textwidth}
		\centering
		
		\input{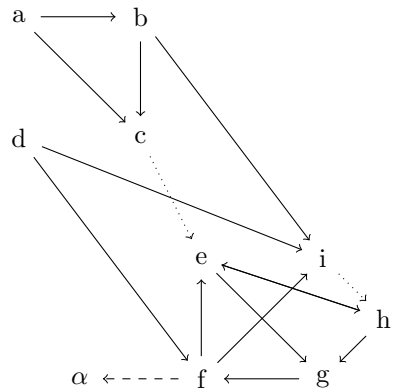}
		
		\caption{The removed arcs from and the added arcs to a reference digraph are indicated by dotted and by dashed lines, respectively.\newline}   \label{fig:example1:modification}
	\end{subfigure}
	\newline
	\begin{subfigure}[b]{0.45\textwidth}
		\centering
		
		\input{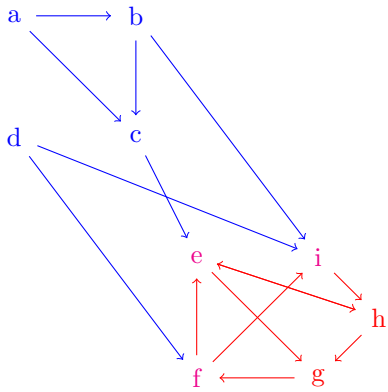}
		
		\caption{The ground and the non-ground of a digraph, if exist, are colored in red and in blue, respectively, except the ground-reciprocal vertices which are colored in magenta.}   \label{fig:example1:digraph_ground}
	\end{subfigure}
	\hspace{0.0cm}
	\begin{subfigure}[b]{0.45\textwidth}
		\centering
		
		\input{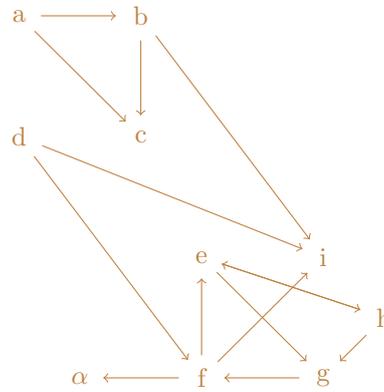}
		
		\caption{A digraph is colored in brown if it does not have a ground. \newline \newline}   \label{fig:example1:digraph_noground}
	\end{subfigure}

	\caption{Visualization of the digraph of Example~\ref{sec:example1} in Figure~\ref{fig:example1:digraph} and the adopted visualization conventions illustrated in Figures~\ref{fig:example1:modification}, \ref{fig:example1:digraph_ground}, and \ref{fig:example1:digraph_noground}.}   \label{fig:example01:1}
\end{figure}

Assume a set of points
\begin{equation}
	\mysetPoints = \{ a, b, c, d, e, f, g, h, i \}\text{,}
\end{equation}
a neighborhod topology
\begin{align}
	\mysetTopology = \{ &(a, \{ b, c \}), (b, \{ c, i \}), (c, \{ e \}), (d, \{ f, i\}),\nonumber \\ 
	&(e, \{ g, h \}), (f, \{ e, i \}), (g, \{ f \}), (h, \{ e, g \}), (i, \{ h \}) \}
\end{align}
on it, and a neighborhood topological space $\mytopologicalSpace$ which is $\mysetPoints$ equipped with $\mysetTopology$. The entities $\mysetPoints$, $\mysetTopology$, and $\mytopologicalSpace$ have the characteristics \#\ref{charsMechantn:discrete} to \#\ref{charsMechantn:comprehensive} of a mechanism listed in Table~\ref{tab:cog_constraints}. Adopting the constitutive postulates (i.e.\ characteristics \#\ref{charsMechantn:discrete}, \#\ref{charsMechantn:restricted}, and \#\ref{charsMechantn:maximalistic}) and characteristics \#\ref{charsMechantn:classic} and \#\ref{charsMechantn:single-leveled}, the neighborhood topological space $\mytopologicalSpace$ becomes a digraph $\mydirectedGraphsym{D}$ which is a set of vertices
\begin{align}
	\mysetVertices{\mydirectedGraphsym{D}} = \{ a, b, c, d, e, f, g, h, i \}
\end{align}
associated with a set of arcs
\begin{align}
	\mysetArcs{\mydirectedGraphsym{D}} = \{ &(a,b), (a,c), (b,c), (b,i), (c,e), (d,f), (d,i), \nonumber \\
	 &(e,g), (e,h), (f,e), (f,i), (g,f), (h,e), (h,g), (i,h) \}\text{.}
\end{align}
Figure~\ref{fig:example01:1} illustrates the digraph $\mydirectedGraphsym{D}$ as well as the visualization conventions adopted for the example problems.
In Figure~\ref{fig:example1:digraph_ground}, a) the digraph $\mydirectedGraphsym{D}$ is the model of a cognition mechanism, b) the sub-digraph composed of vertices e, f, g, h, and i and all the arcs between them is the ground of $\mydirectedGraphsym{D}$, termed $\myfuncground{\mydirectedGraphsym{D}}$, which is the model of the cognition mechanism's self, c) the sub-digraph composed of vertices a, b, c, d, e, f, and i and all the arcs between them is the non-ground of $\mydirectedGraphsym{D}$, termed $\myfuncnonground{\mydirectedGraphsym{D}}$, which is the model of the cognition mechanism's non-self, and d) the set of vertices e, f, and i are the set of ground-reciprocal vertices of $\mydirectedGraphsym{D}$, termed $\myfuncreciprocalground{\mydirectedGraphsym{D}}$, which is the model of the group of self-mutual nodes of the cognition mechanism.\footnote{The digraph is drawn such that the ground is visually distant from and in a different color from the non-ground for the sake of visually better identification of them.}).

\subsection{Presentation of characterization and formization of mechanisms}   \label{sec:exampl_pres_charform_mechs}
This example is designated to provide a presentation type of characterization and formization of mechanisms to the reader.

\begin{figure}
	\centering
		
	\input{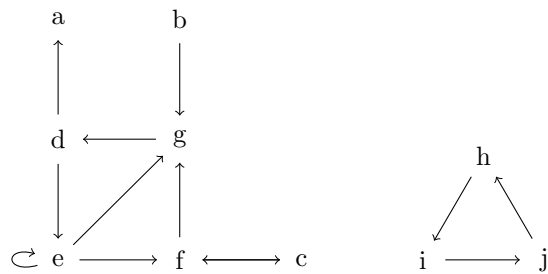}
	
	\caption{Visualizations of the digraph in Example~\ref{sec:example:mechProp_mech}.}   \label{fig:example2:digraph}
\end{figure}

Assume a digraph $\mydirectedGraphsym{D}$ which is a set of vertices 
\begin{align}
	\mysetVertices{\mydirectedGraphsym{D}} = \{ a, b, c, d, e, f, g, h, i, j \}
\end{align}
associated with a set of arcs
\begin{align}
	\mysetArcs{\mydirectedGraphsym{D}} = \{ (b,g), (c,f), (d,a), (d,e), (e,e), (e,f), (e,g), (f,c), (f,g), (g,d), (h,i), (i,j), (j,h) \}
\end{align}
and which is illustrated in Figure~\ref{fig:example2:digraph}.
All paths and cycles of $\mydirectedGraphsym{D}$ are listed in Table~\ref{tab:exmple2:paths_cycles}. All walks of this table have the characteristics \#\ref{charsMechantn:actual} to \#\ref{charsMechantn:exhaustive} of a mechanism listed in Table~\ref{tab:cog_constraints} with an exception for the listed cycles which do not satisfy \#\ref{charsMechantn:diversal}.
\begin{table}
	\centering
	\caption{All paths and cycles of the digraph $\mydirectedGraphsym{D}$ in Example~\ref{sec:example:mechProp_mech}.}   \label{tab:exmple2:paths_cycles}
	\begin{tabular}{l}
		\hline
		\textbf{paths}\\
		\hline
		$b \, g \, d \, a$,
		$b \, g \, d \, e \, f \, c$, 
		$b \, g \, d$, 
		$b \, g \, d \, e$, 
		$b \, g \, d \, e \, f$, 
		$b \, g$\\
		$c \, f \, g \, d \, a$, 
		$c \, f \, g \, d$, 
		$c \, f \, g \, d \, e$, 
		$c \, f$, 
		$c \, f \, g$\\
		$d \, a$, 
		$d \, e \, f \, c$, 
		$d \, e$, 
		$d \, e \, f$, 
		$d \, e \, f \, g$, 
		$d \, e \, g$\\
		$e \, f \, g \, d \, a$, 
		$e \, g \, d \, a$, 
		$e \, f \, c$, 
		$e \, f \, g \, d$, 
		$e \, g \, d$, 
		$e \, e$, 
		$e \, f$, 
		$e \, f \, g$, 
		$e \, g$\\
		$f \, g \, d \, a$, 
		$f \, c$, 
		$f \, g \, d$, 
		$f \, g \, d \, e$, 
		$f \, g$\\
		$g \, d \, a$, 
		$g \, d \, e \, f \, c$, 
		$g \, d$, 
		$g \, d \, e$, 
		$g \, d \, e \, f$\\
		$h \, i$,
		$h \, i \, j$\\
		$i \, j \, h$,
		$i \, j$\\
		$j \, h$,
		$j \, h \, i$\\
		\hline
		\hline
		\textbf{cycles}\\
		\hline
		$c \, f \, c$\\
		$d \, e \, f \, g \, d$, 
		$d \, e \, g \, d$\\
		$e \, e$, 
		$e \, f \, g \, d \, e$, 
		$e \, g \, d \, e$\\
		$f \, c \, f$, 
		$f \, g \, d \, e \, f$\\
		$g \, d \, e \, f \, g$, 
		$g \, d \, e \, g$\\
		$h \, i \, j \, h$\\
		$i \, j \, h \, i$\\
		$j \, h \, i \, j$\\
		\hline
	\end{tabular}
\end{table}

\begin{figure}
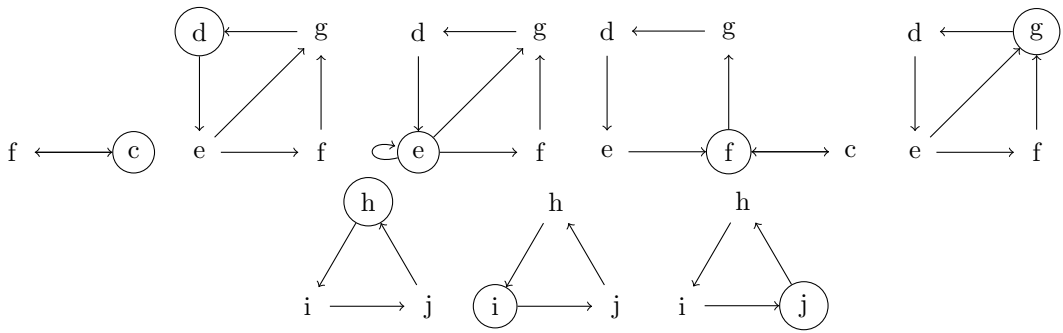

	\centering
	
	\input{exmpl2_digraph_unit_c.tikz}
	\input{exmpl2_digraph_unit_d.tikz}
	\input{exmpl2_digraph_unit_e.tikz}
	\input{exmpl2_digraph_unit_f.tikz}
	\input{exmpl2_digraph_unit_g.tikz}\newline%
	\input{exmpl2_digraph_unit_h.tikz}
	\input{exmpl2_digraph_unit_i.tikz}
	\input{exmpl2_digraph_unit_j.tikz}
	
	\caption{Visualizations of all-cyclic sub-digraphs of the digraph $\mydirectedGraphsym{D}$ in Example~\ref{sec:exampl_pres_charform_mechs}. An all-cyclic sub-digraph over vertex $x$ is illustrated by the sub-digraph with a circle around vertex $x$ (where $x$ ranges over all vertices in $\mysetVertices{\mydirectedGraphsym{D}}$).}   \label{fig:example2:allcyclicsubdigraphs}
\end{figure}

\begin{figure}
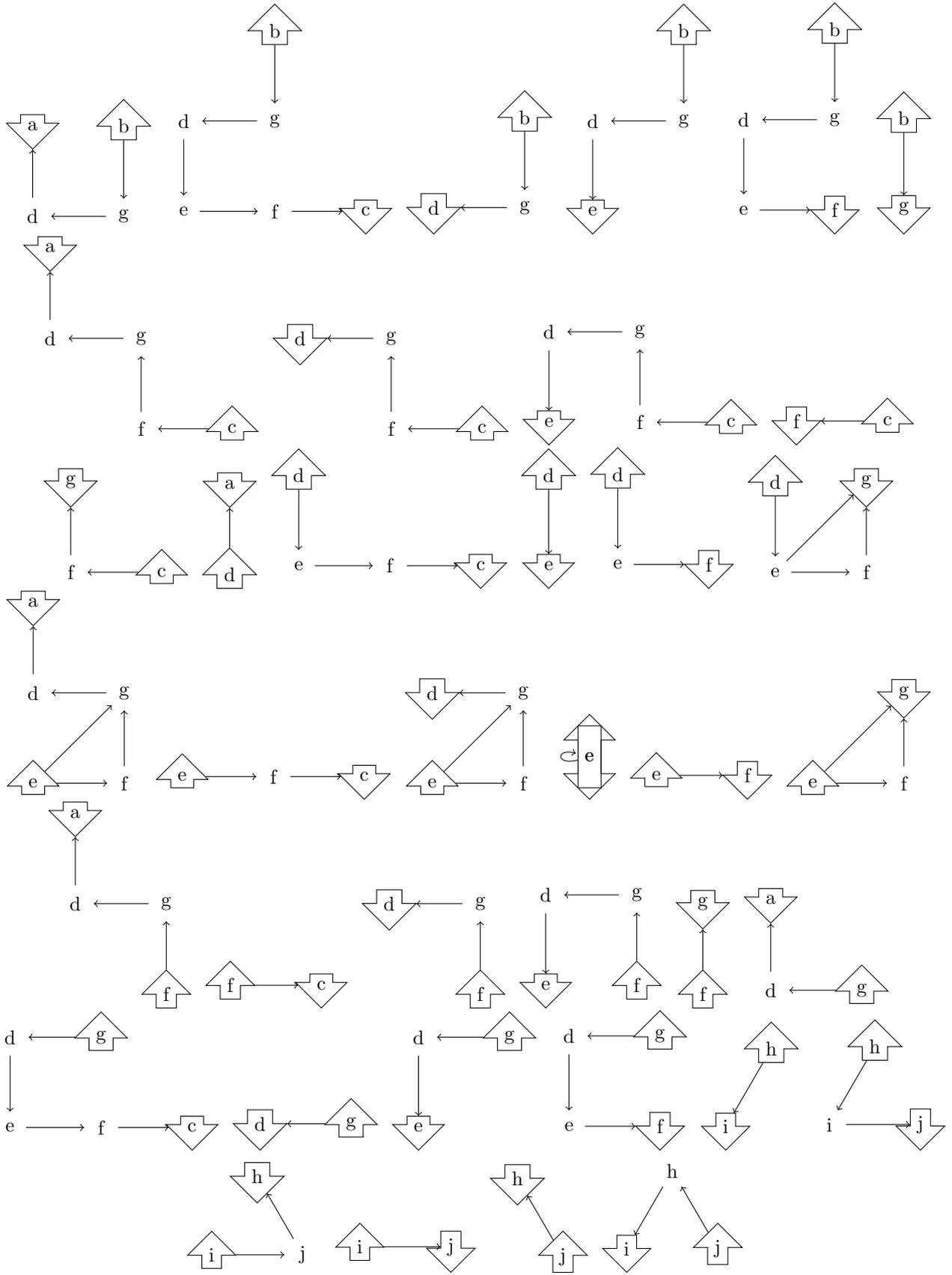

	\centering
	
	\input{exmpl2_digraph_uniter_ba.tikz}
	\input{exmpl2_digraph_uniter_bc.tikz}
	\input{exmpl2_digraph_uniter_bd.tikz}
	\input{exmpl2_digraph_uniter_be.tikz}
	\input{exmpl2_digraph_uniter_bf.tikz}
	\input{exmpl2_digraph_uniter_bg.tikz} %
	\input{exmpl2_digraph_uniter_ca.tikz}
	\input{exmpl2_digraph_uniter_cd.tikz}
	\input{exmpl2_digraph_uniter_ce.tikz}
	\input{exmpl2_digraph_uniter_cf.tikz}
	\input{exmpl2_digraph_uniter_cg.tikz} %
	\input{exmpl2_digraph_uniter_da.tikz}
	\input{exmpl2_digraph_uniter_dc.tikz}
	\input{exmpl2_digraph_uniter_de.tikz}
	\input{exmpl2_digraph_uniter_df.tikz}
	\input{exmpl2_digraph_uniter_dg.tikz} %
	\input{exmpl2_digraph_uniter_ea.tikz}
	\input{exmpl2_digraph_uniter_ec.tikz}
	\input{exmpl2_digraph_uniter_ed.tikz}
	\input{exmpl2_digraph_uniter_ee.tikz}
	\input{exmpl2_digraph_uniter_ef.tikz}
	\input{exmpl2_digraph_uniter_eg.tikz} %
	\input{exmpl2_digraph_uniter_fa.tikz}
	\input{exmpl2_digraph_uniter_fc.tikz}
	\input{exmpl2_digraph_uniter_fd.tikz}
	\input{exmpl2_digraph_uniter_fe.tikz}
	\input{exmpl2_digraph_uniter_fg.tikz} %
	\input{exmpl2_digraph_uniter_ga.tikz}
	\input{exmpl2_digraph_uniter_gc.tikz}
	\input{exmpl2_digraph_uniter_gd.tikz}
	\input{exmpl2_digraph_uniter_ge.tikz}
	\input{exmpl2_digraph_uniter_gf.tikz} %
	\input{exmpl2_digraph_uniter_hi.tikz}
	\input{exmpl2_digraph_uniter_hj.tikz}
	\input{exmpl2_digraph_uniter_ih.tikz}
	\input{exmpl2_digraph_uniter_ij.tikz}
	\input{exmpl2_digraph_uniter_jh.tikz}
	\input{exmpl2_digraph_uniter_ji.tikz} %
	
	\caption{Visualizations of all-pathic sub-digraphs of the digraph $\mydirectedGraphsym{D}$ in Example~\ref{sec:exampl_pres_charform_mechs}. An all-pathic sub-digraph from vertex $x$ to vertex $y$ is illustrated by the sub-digraph with an upward and downward arrow around $x$ and $y$, respectively (where $x$ and $y$ range over all vertices in $\mysetVertices{\mydirectedGraphsym{D}}$).}   \label{fig:example2:allpathicsubdigraphs}
\end{figure}

Assume a mechanism $\mymechsym$. Its network $\myconstituentsym$ and its intramission capacity $\myemergentsym$ are modeled by the digraph $\mydirectedGraphsym{D}$ and the process of a walk in it, respectively. For the mechanistic characterization of $\mymechsym$, all units and/or uniters of $\myconstituentsym$ should be found. In this regard and in mathematical terms, according to Table~\ref{tab:exmple2:paths_cycles}, every all-cyclic sub-digraph and every all-pathic sub-digraph of $\mydirectedGraphsym{D}$ are visualized in Figures~\ref{fig:example2:allcyclicsubdigraphs} and \ref{fig:example2:allpathicsubdigraphs}, respectively. Recognition of all-cyclic and all-pathic sub-digraphs in these figures is straightforward; however, it may be helpful to remind that, a) in an all-cyclic sub-digraph over vertex $x$, all other vertices of the sub-digraph are connected to and from vertex $x$ and b) in an all-pathic sub-digraph from vertex $x$ to vertex $y$, all other vertices of the sub-digraph are connected from vertex $x$ and to vertex $y$. Note that, as also seen from Figures~\ref{fig:example2:allcyclicsubdigraphs} and \ref{fig:example2:allpathicsubdigraphs}, $\mydirectedGraphsym{D}$ cannot be characterized symbolistically while can be characterized connectionistically.

Assume that $\mymechsym$, $\myconstituentsym$, and $\myemergentsym$ are in form $\mathrm{base}$ and name them $\mymechsym_\mathrm{base}$, $\myconstituentsym_\mathrm{base}$, and $\myemergentsym_\mathrm{base}$, respectively. An iso-mechanation from form `$\mathrm{base}$' to `$\mathrm{iso}$' leads to an iso-mechanism $\mymechsym_\mathrm{iso}$ with constituents $\myconstituentsym_\mathrm{iso}$ and an intramission capacity $\myemergentsym_\mathrm{iso}$. A higher-order mechanistic formization of $\mymechsym_\mathrm{base}$ leads to $\mymechsym_\mathrm{iso}$ whose $\myconstituentsym_\mathrm{iso}$'s a) model has ten vertices, named $o1$, $o2$, \ldots, $o10$\footnote{Note that the vertices of $\myconstituentsym_\mathrm{iso}$'s model does not have real labels and thus the indicated labels of these vertices are imaginarily assigned in order to more easily deal with them.}, which are completely connected to each other and b) model's vertices and arcs have property measures of Table~\ref{tab:exmple2:formiz:nodaldilinkal_props}.
\begin{table}
	\centering
	
	\caption{The number of paths between vertices of, the number of cycles over vertices of, and the number of vertices adjacent from and to vertices of $\myconstituentsym_\mathrm{iso}$'s model in Example~\ref{sec:exampl_pres_charform_mechs}. Each path is considered from a vertex listed in the left column to a vertex listed in the top row, and the diagonal elements of the table refer to loops over vertices. Each loop over a vertex is counted in the corresponding row for the number of cycles over that vertex as well as in both of the number of vertices adjacent from and the number of vertices adjacent to that vertex.} \label{tab:exmple2:formiz:nodaldilinkal_props}
	
	\begin{tabular}{| c || c | c | c | c | c | c | c | c | c | c |}
		\multicolumn{11}{c}{\textbf{number of paths}}\\
		\hline
		$\mypathsymne$ & o1 & o2 & o3 & o4 & o5 & o6 & o7 & o8 & o9 & o10\\
		\hline
		\hline
		o1 & 0 & 0 & 0 & 0 & 0 & 0 & 0 & 0 & 0 & 0\\
		\hline
		o2 & 1 & 0 & 1 & 1 & 1 & 1 & 1 & 0 & 0 & 0\\
		\hline
		o3 & 1 & 0 & 0 & 1 & 1 & 1 & 1 & 0 & 0 & 0\\
		\hline
		o4 & 1 & 0 & 1 & 0 & 1 & 1 & 2 & 0 & 0 & 0\\
		\hline
		o5 & 2 & 0 & 1 & 2 & 1 & 1 & 2 & 0 & 0 & 0\\
		\hline
		o6 & 1 & 0 & 1 & 1 & 1 & 0 & 1 & 0 & 0 & 0\\
		\hline
		o7 & 1 & 0 & 1 & 1 & 1 & 1 & 0 & 0 & 0 & 0\\
		\hline
		o8 & 0 & 0 & 0 & 0 & 0 & 0 & 0 & 0 & 1 & 1\\
		\hline
		o9 & 0 & 0 & 0 & 0 & 0 & 0 & 0 & 1 & 0 & 1\\
		\hline
		o10 & 0 & 0 & 0 & 0 & 0 & 0 & 0 & 1 & 1 & 0\\
		\hline
		\multicolumn{11}{c}{\textbf{number of cycles}}\\
		\hline
		over ($\circlearrowright \circ$) & 0 & 0 & 1 & 2 & 3 & 2 & 2 & 1 & 1 & 1\\
		\hline
		\multicolumn{11}{c}{\textbf{number of arcs}}\\
		\hline
		from ($\circ \rightarrow$) & 0 & 1 & 1 & 2 & 3 & 2 & 1 & 1 & 1 & 1\\
		\hline
		to ($\rightarrow \circ$) & 1 & 0 & 1 & 1 & 2 & 2 & 3 & 1 & 1 & 1\\
		\hline
	\end{tabular}
\end{table}
According to formization modes stated in Section~\ref{sec:metainfra_mechs:mech_formiz}, it is sufficient for a higher order mechanistic formization to determine either a) `the number of paths between' and `the number of cycles over' or b) `the number of paths between' and `the numbers of vertices adjacent from and to' the vertices of $\myconstituentsym_\mathrm{iso}$'s model. Nonetheless, they all are brought in Table~\ref{tab:exmple2:formiz:nodaldilinkal_props} so that the reader becomes acquainted with a complete presentation type of the formization of a mechanism. In addition, $\myemergentsym_\mathrm{iso}$ which is a process that determines real labels of nodes can be rigorously asserted by a mathematical algorithm which is not specified here for the sake of brevity. It suffices here to know that, because nodes of $\myconstituentsym_\mathrm{iso}$ do not have real labels, the formization table of $\mymechsym_\mathrm{iso}$ can be reordered such that if the elements of $i$th and $j$th columns related to all property measures are swapped, the elements of $i$th and $j$th rows related to the number of paths between nodes must also be swapped. Thus, $\mymechsym_\mathrm{iso}$'s formization table may have several appearances (according to the point mentioned) all of which are equivalently formizing $\mymechsym_\mathrm{iso}$.

\subsection{Mechanistic analysis of mechanisms}   \label{sec:example:mechProp_mech}
This analyzes mechanisms through the mechanistic characterization and formization proposed in Section~\ref{sec:metainfra_mechs:mech_formiz}.

\subsubsection{Mechanistic characterization}   \label{sec:example:mechProp_mech:charmodes}
\begin{figure}
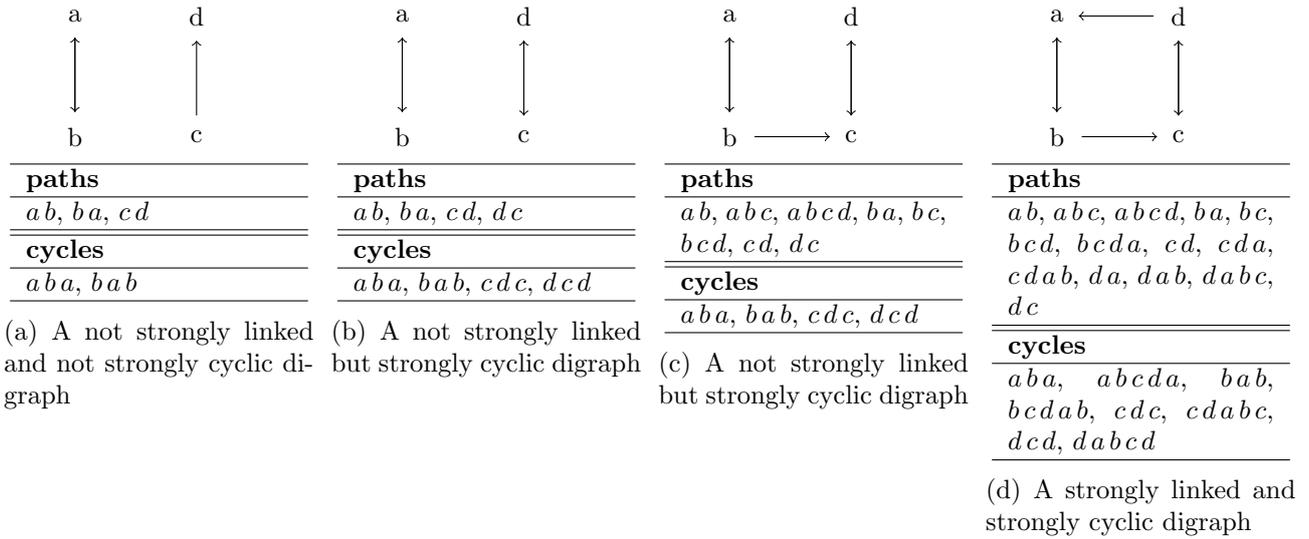

	\centering
	
	\begin{subfigure}[t]{0.24\textwidth}
		\centering
		
		\input{exmpl2_digraph_notlinked_notcyclic.tikz}\newline%
		\small
		\begin{tabular}{p{3.5cm}}
			\hline
			\textbf{paths}\\
			\hline
			$a \, b$, 
			$b \, a$, 
			$c \, d$\\
			\hline
			\hline
			\textbf{cycles}\\
			\hline
			$a \, b \, a$,
			$b \, a \, b$\\
			\hline
		\end{tabular}
		
		\caption{A not strongly linked and not strongly cyclic digraph}   \label{fig:example2:char_modes:nlk_ncy}
	\end{subfigure}
	\hspace{0.0cm}
	\begin{subfigure}[t]{0.24\textwidth}
		\centering
		
		\input{exmpl2_digraph_notlinked_cyclic_I.tikz}\newline%
		\small
		\begin{tabular}{p{3.5cm}}
			\hline
			\textbf{paths}\\
			\hline
			$a \, b$, 
			$b \, a$, 
			$c \, d$,
			$d \, c$\\
			\hline
			\hline
			\textbf{cycles}\\
			\hline
			$a \, b \, a$,
			$b \, a \, b$,
			$c \, d \, c$,
			$d \, c \, d$\\
			\hline
		\end{tabular}
		
		\caption{A not strongly linked but strongly cyclic digraph}   \label{fig:example2:char_modes:nlk_cy_I}
	\end{subfigure}
	\hspace{0.0cm}
	\begin{subfigure}[t]{0.24\textwidth}
		\centering
		
		\input{exmpl2_digraph_notlinked_cyclic_II.tikz}\newline%
		\small
		\begin{tabular}{p{3.5cm}}
			\hline
			\textbf{paths}\\
			\hline
			$a \, b$, $a \, b \, c$, $a \, b \, c \, d$, 
			$b \, a$, $b \, c$, $b \, c \, d$,  
			$c \, d$,
			$d \, c$\\
			\hline
			\hline
			\textbf{cycles}\\
			\hline
			$a \, b \, a$,
			$b \, a \, b$,
			$c \, d \, c$,
			$d \, c \, d$\\
			\hline
		\end{tabular}
		
		\caption{A not strongly linked but strongly cyclic digraph}   \label{fig:example2:char_modes:nlk_cy_II}
	\end{subfigure}
	\hspace{0.0cm}
	\begin{subfigure}[t]{0.24\textwidth}
		\centering
		
		\input{exmpl2_digraph_linked_cyclic.tikz}\newline%
		\small
		\begin{tabular}{p{3.5cm}}
			\hline
			\textbf{paths}\\
			\hline
			$a \, b$, $a \, b \, c$, $a \, b \, c \, d$, 
			$b \, a$, $b \, c$, $b \, c \, d$, $b \, c \, d \, a$,
			$c \, d$, $c \, d \, a$, $c \, d \, a \, b$,
			$d \, a$, $d \, a \, b$, $d \, a \, b \, c$, $d \, c$\\
			\hline
			\hline
			\textbf{cycles}\\
			\hline
			$a \, b \, a$,
			$a \, b \, c \, d \, a$,
			$b \, a \, b$,
			$b \, c \, d \, a \, b$,
			$c \, d \, c$,
			$c \, d \, a \, b \, c$,
			$d \, c \, d$,
			$d \, a \, b \, c \, d$\\
			\hline
		\end{tabular}
		
		\caption{A strongly linked and strongly cyclic digraph}   \label{fig:example2:char_modes:lk_cy}
	\end{subfigure}
	
	\caption{Visualizations of the digraphs of Example~\ref{sec:example:mechProp_mech:charmodes} and their paths and cycles.}   \label{fig:example02:char_modes}
\end{figure}

Assume the digraphs illustrated in Figure~\ref{fig:example02:char_modes}.
According to their cycles, the digraph~\ref{fig:example2:char_modes:nlk_ncy} cannot be (totally) characterized symbolistically because the arc $(c, d)$ will not be included. The situation is similar for the digraph~\ref{fig:example2:char_modes:nlk_cy_II} and its arc $(b, c)$. The digraphs~\ref{fig:example2:char_modes:nlk_cy_I} and \ref{fig:example2:char_modes:lk_cy} can be (totally) characterized symbolistically. However, by comparing the digraphs~\ref{fig:example2:char_modes:nlk_cy_I} and \ref{fig:example2:char_modes:nlk_cy_II}, it is concluded that being strongly cyclic is not a sufficient condition for a digraph to be able to symbolistically characterize it. In addition, according to their paths, all these digraphs can be (totally) characterized connectionistically as well as hybridistically.

\subsubsection{Mechanistic formization}   \label{sec:example:mechProp_mech:formizmodes}

\paragraph{Case 1}
Assume the digraphs illustrated in Figure~\ref{fig:example03:formiz_modes}.
\begin{figure}
	\centering
	
	\begin{subfigure}[t]{0.45\textwidth}
		\centering
		
		\input{exmpl3_digraph_formiz_var1.tikz}\\%
		\scriptsize
		\begin{tabular}{| c || c | c | c | c | c | c |}
			\multicolumn{7}{c}{\textbf{number of paths}}\\
			\hline
			$\mypathsymne$ & o1 & o2 & o3 & o4 & o5 & o6\\
			\hline
			\hline
			o1 & 0 & 2 & 2 & 0 & 0 & 0\\
			\hline
			o2 & 2 & 0 & 2 & 0 & 0 & 0\\
			\hline
			o3 & 2 & 2 & 0 & 0 & 0 & 0\\
			\hline
			o4 & 0 & 0 & 0 & 0 & 1 & 1\\
			\hline
			o5 & 0 & 0 & 0 & 1 & 0 & 1\\
			\hline
			o6 & 0 & 0 & 0 & 1 & 1 & 0\\
			\hline
			\multicolumn{7}{c}{\textbf{number of cycles}}\\
			\hline
			$\circlearrowright \circ$ & 4 & 4 & 4 & 1 & 1 & 1\\
			\hline
			\multicolumn{7}{c}{\textbf{number of arcs}}\\
			\hline
			$\circ \rightarrow$ & 2 & 2 & 2 & 1 & 1 & 1\\
			\hline
			$\rightarrow \circ$ & 2 & 2 & 2 & 1 & 1 & 1\\
			\hline
		\end{tabular}
		
		\caption{Digraph $\mydirectedGraphsym{D}1$}   \label{fig:example2:formiz_modes:digraph_var1}
	\end{subfigure}\\%
	\begin{subfigure}[t]{0.45\textwidth}
		\centering
		
		\input{exmpl3_digraph_formiz_var2.tikz}\\%
		\scriptsize
		\begin{tabular}{| c || c | c | c | c | c | c |}
			\multicolumn{7}{c}{\textbf{number of paths}}\\
			\hline
			$\mypathsymne$ & o1 & o2 & o3 & o4 & o5 & o6\\
			\hline
			\hline
			o1 & 0 & 2 & 2 & 2 & 2 & 2\\
			\hline
			o2 & 2 & 0 & 2 & 2 & 2 & 2\\
			\hline
			o3 & 2 & 2 & 0 & 1 & 1 & 1\\
			\hline
			o4 & 2 & 1 & 2 & 0 & 1 & 1\\
			\hline
			o5 & 2 & 1 & 2 & 2 & 0 & 1\\
			\hline
			o6 & 2 & 1 & 2 & 2 & 2 & 0\\
			\hline
			\multicolumn{7}{c}{\textbf{number of cycles}}\\
			\hline
			$\circlearrowright \circ$ & 4 & 4 & 4 & 2 & 2 & 2\\
			\hline
			\multicolumn{7}{c}{\textbf{number of arcs}}\\
			\hline
			$\circ \rightarrow$ & 2 & 2 & 2 & 1 & 1 & 1\\
			\hline
			$\rightarrow \circ$ & 2 & 2 & 2 & 1 & 1 & 1\\
			\hline
		\end{tabular}
		
		\caption{Digraph $\mydirectedGraphsym{D}2$}   \label{fig:example2:formiz_modes:digraph_var2}
	\end{subfigure}
	\hspace{0.0cm}
	\begin{subfigure}[t]{0.45\textwidth}
		\centering
		
		\input{exmpl3_digraph_formiz_var3.tikz}\\%
		\scriptsize
		\begin{tabular}{| c || c | c | c | c | c | c |}
			\multicolumn{7}{c}{\textbf{number of paths}}\\
			\hline
			$\mypathsymne$ & o1 & o2 & o3 & o4 & o5 & o6\\
			\hline
			\hline
			o1 & 0 & 1 & 1 & 1 & 1 & 1\\
			\hline
			o2 & 1 & 0 & 1 & 1 & 1 & 1\\
			\hline
			o3 & 1 & 1 & 0 & 1 & 1 & 1\\
			\hline
			o4 & 1 & 1 & 1 & 0 & 1 & 1\\
			\hline
			o5 & 1 & 1 & 1 & 1 & 0 & 1\\
			\hline
			o6 & 1 & 1 & 1 & 1 & 1 & 0\\
			\hline
			\multicolumn{7}{c}{\textbf{number of cycles}}\\
			\hline
			$\circlearrowright \circ$ & 1 & 1 & 1 & 1 & 1 & 1\\
			\hline
			\multicolumn{7}{c}{\textbf{number of arcs}}\\
			\hline
			$\circ \rightarrow$ & 1 & 1 & 1 & 1 & 1 & 1\\
			\hline
			$\rightarrow \circ$ & 1 & 1 & 1 & 1 & 1 & 1\\
			\hline
		\end{tabular}
		
		\caption{Digraph $\mydirectedGraphsym{D}3$}   \label{fig:example2:formiz_modes:digraph_var3}
	\end{subfigure}\\%
	\begin{subfigure}[t]{0.45\textwidth}
		\centering
		
		\input{exmpl3_digraph_formiz_var4.tikz}\\%
		\scriptsize
		\begin{tabular}{| c || c | c | c | c | c | c |}
			\multicolumn{7}{c}{\textbf{number of paths}}\\
			\hline
			$\mypathsymne$ & o1 & o2 & o3 & o4 & o5 & o6\\
			\hline
			\hline
			o1 & 0 & 2 & 2 & 5 & 5 & 5\\
			\hline
			o2 & 2 & 0 & 2 & 5 & 5 & 5\\
			\hline
			o3 & 2 & 2 & 0 & 5 & 5 & 5\\
			\hline
			o4 & 0 & 0 & 0 & 0 & 1 & 1\\
			\hline
			o5 & 0 & 0 & 0 & 1 & 0 & 1\\
			\hline
			o6 & 0 & 0 & 0 & 1 & 1 & 0\\
			\hline
			\multicolumn{7}{c}{\textbf{number of cycles}}\\
			\hline
			$\circlearrowright \circ$ & 4 & 4 & 4 & 1 & 1 & 1\\
			\hline
			\multicolumn{7}{c}{\textbf{number of arcs}}\\
			\hline
			$\circ \rightarrow$ & 3 & 3 & 3 & 1 & 1 & 1\\
			\hline
			$\rightarrow \circ$ & 2 & 2 & 2 & 2 & 2 & 2\\
			\hline
		\end{tabular}
		
		\caption{Digraph $\mydirectedGraphsym{D}4$}   \label{fig:example2:formiz_modes:digraph_var4}
	\end{subfigure}
	\hspace{0.0cm}
	\begin{subfigure}[t]{0.45\textwidth}
		\centering
		
		\input{exmpl3_digraph_formiz_var5.tikz}\\%
		\scriptsize
		\begin{tabular}{| c || c | c | c | c | c | c |}
			\multicolumn{7}{c}{\textbf{number of paths}}\\
			\hline
			$\mypathsymne$ & o1 & o2 & o3 & o4 & o5 & o6\\
			\hline
			\hline
			o1 & 0 & 1 & 1 & 1 & 1 & 1\\
			\hline
			o2 & 1 & 0 & 1 & 1 & 1 & 1\\
			\hline
			o3 & 1 & 1 & 0 & 1 & 1 & 1\\
			\hline
			o4 & 1 & 1 & 1 & 0 & 1 & 1\\
			\hline
			o5 & 1 & 1 & 1 & 1 & 0 & 1\\
			\hline
			o6 & 1 & 1 & 1 & 1 & 1 & 0\\
			\hline
			\multicolumn{7}{c}{\textbf{number of cycles}}\\
			\hline
			$\circlearrowright \circ$ & 2 & 2 & 2 & 1 & 1 & 1\\
			\hline
			\multicolumn{7}{c}{\textbf{number of arcs}}\\
			\hline
			$\circ \rightarrow$ & 2 & 2 & 2 & 1 & 1 & 1\\
			\hline
			$\rightarrow \circ$ & 2 & 2 & 2 & 1 & 1 & 1\\
			\hline
		\end{tabular}
		
		\caption{Digraph $\mydirectedGraphsym{D}5$}   \label{fig:example2:formiz_modes:digraph_var5}
	\end{subfigure}
	
	\caption{Visualizations of the digraphs of Case 1 of Example~\ref{sec:example:mechProp_mech:formizmodes} and the information for their formizations.}   \label{fig:example03:formiz_modes}
\end{figure}
It is seen from the formization tables of the digraphs that
\begin{itemize}
	\item   $\mydirectedGraphsym{D}1$, $\mydirectedGraphsym{D}2$, and $\mydirectedGraphsym{D}5$ have the same number of vertices adjacent from and to their vertices.
	
	\item   $\mydirectedGraphsym{D}1$ and $\mydirectedGraphsym{D}4$ have the same number of cycles over their vertices.
	
	\item   $\mydirectedGraphsym{D}3$ and $\mydirectedGraphsym{D}5$ have the same number of paths between their vertices.
	
	\item   No two digraphs have the same formization in either of the formization modes.
\end{itemize}
From these statements, one concludes that no single measure of the measures can (totally) formize a digraph in general.

\begin{figure}
	\centering
	
	\begin{subfigure}[t]{0.45\textwidth}
		\centering
		
		\input{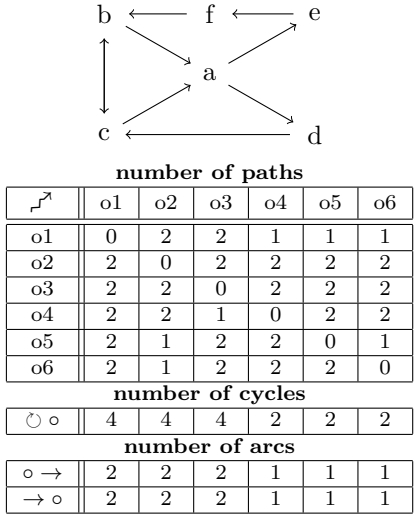}\\%
		\scriptsize
		\begin{tabular}{| c || c | c | c | c | c | c |}
			\multicolumn{7}{c}{\textbf{number of paths}}\\
			\hline
			$\mypathsymne$ & o1 & o2 & o3 & o4 & o5 & o6\\
			\hline
			\hline
			o1 & 0 & 2 & 2 & 1 & 1 & 1\\
			\hline
			o2 & 2 & 0 & 2 & 2 & 2 & 2\\
			\hline
			o3 & 2 & 2 & 0 & 2 & 2 & 2\\
			\hline
			o4 & 2 & 2 & 1 & 0 & 2 & 2\\
			\hline
			o5 & 2 & 1 & 2 & 2 & 0 & 1\\
			\hline
			o6 & 2 & 1 & 2 & 2 & 2 & 0\\
			\hline
			\multicolumn{7}{c}{\textbf{number of cycles}}\\
			\hline
			$\circlearrowright \circ$ & 4 & 4 & 4 & 2 & 2 & 2\\
			\hline
			\multicolumn{7}{c}{\textbf{number of arcs}}\\
			\hline
			$\circ \rightarrow$ & 2 & 2 & 2 & 1 & 1 & 1\\
			\hline
			$\rightarrow \circ$ & 2 & 2 & 2 & 1 & 1 & 1\\
			\hline
		\end{tabular}
		
		\caption{Digraph $\mydirectedGraphsym{D}6_{I}$}   \label{fig:example2:formiz_modes:digraph_var6_1}
	\end{subfigure}
	\hspace{0.0cm}
	\begin{subfigure}[t]{0.45\textwidth}
		\centering
		
		\input{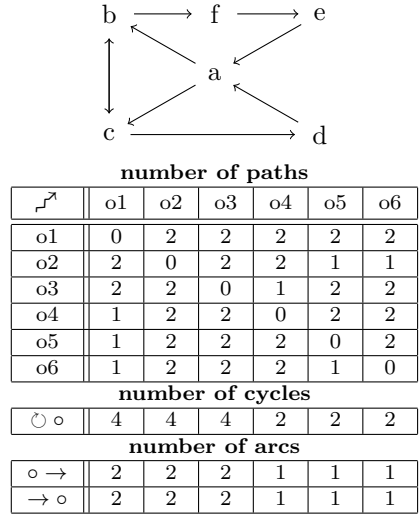}\\%
		\scriptsize
		\begin{tabular}{| c || c | c | c | c | c | c |}
			\multicolumn{7}{c}{\textbf{number of paths}}\\
			\hline
			$\mypathsymne$ & o1 & o2 & o3 & o4 & o5 & o6\\
			\hline
			\hline
			o1 & 0 & 2 & 2 & 2 & 2 & 2\\
			\hline
			o2 & 2 & 0 & 2 & 2 & 1 & 1\\
			\hline
			o3 & 2 & 2 & 0 & 1 & 2 & 2\\
			\hline
			o4 & 1 & 2 & 2 & 0 & 2 & 2\\
			\hline
			o5 & 1 & 2 & 2 & 2 & 0 & 2\\
			\hline
			o6 & 1 & 2 & 2 & 2 & 1 & 0\\
			\hline
			\multicolumn{7}{c}{\textbf{number of cycles}}\\
			\hline
			$\circlearrowright \circ$ & 4 & 4 & 4 & 2 & 2 & 2\\
			\hline
			\multicolumn{7}{c}{\textbf{number of arcs}}\\
			\hline
			$\circ \rightarrow$ & 2 & 2 & 2 & 1 & 1 & 1\\
			\hline
			$\rightarrow \circ$ & 2 & 2 & 2 & 1 & 1 & 1\\
			\hline
		\end{tabular}
		
		\caption{Digraph $\mydirectedGraphsym{D}6_{II}$}   \label{fig:example2:formiz_modes:digraph_var6_2}
	\end{subfigure}\\%
	\begin{subfigure}[t]{0.45\textwidth}
		\centering
		
		\input{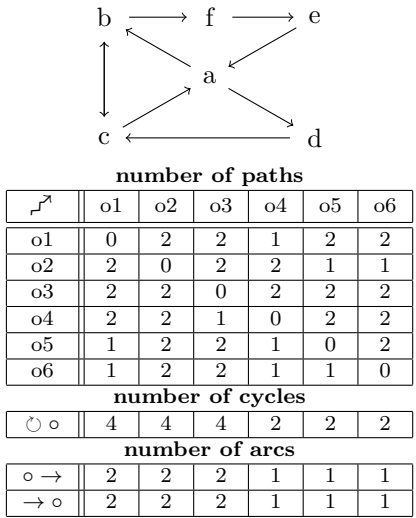}\\%
		\scriptsize
		\begin{tabular}{| c || c | c | c | c | c | c |}
			\multicolumn{7}{c}{\textbf{number of paths}}\\
			\hline
			$\mypathsymne$ & o1 & o2 & o3 & o4 & o5 & o6\\
			\hline
			\hline
			o1 & 0 & 2 & 2 & 1 & 2 & 2\\
			\hline
			o2 & 2 & 0 & 2 & 2 & 1 & 1\\
			\hline
			o3 & 2 & 2 & 0 & 2 & 2 & 2\\
			\hline
			o4 & 2 & 2 & 1 & 0 & 2 & 2\\
			\hline
			o5 & 1 & 2 & 2 & 1 & 0 & 2\\
			\hline
			o6 & 1 & 2 & 2 & 1 & 1 & 0\\
			\hline
			\multicolumn{7}{c}{\textbf{number of cycles}}\\
			\hline
			$\circlearrowright \circ$ & 4 & 4 & 4 & 2 & 2 & 2\\
			\hline
			\multicolumn{7}{c}{\textbf{number of arcs}}\\
			\hline
			$\circ \rightarrow$ & 2 & 2 & 2 & 1 & 1 & 1\\
			\hline
			$\rightarrow \circ$ & 2 & 2 & 2 & 1 & 1 & 1\\
			\hline
		\end{tabular}
		
		\caption{Digraph $\mydirectedGraphsym{D}6_{III}$}   \label{fig:example2:formiz_modes:digraph_var6_3}
	\end{subfigure}
	\hspace{0.0cm}
	\begin{subfigure}[t]{0.45\textwidth}
		\centering
		
		\input{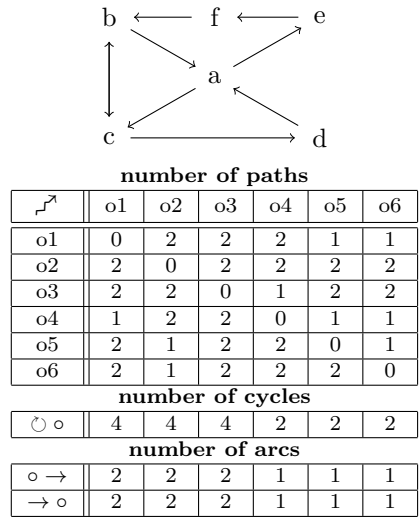}\\%
		\scriptsize
		\begin{tabular}{| c || c | c | c | c | c | c |}
			\multicolumn{7}{c}{\textbf{number of paths}}\\
			\hline
			$\mypathsymne$ & o1 & o2 & o3 & o4 & o5 & o6\\
			\hline
			\hline
			o1 & 0 & 2 & 2 & 2 & 1 & 1\\
			\hline
			o2 & 2 & 0 & 2 & 2 & 2 & 2\\
			\hline
			o3 & 2 & 2 & 0 & 1 & 2 & 2\\
			\hline
			o4 & 1 & 2 & 2 & 0 & 1 & 1\\
			\hline
			o5 & 2 & 1 & 2 & 2 & 0 & 1\\
			\hline
			o6 & 2 & 1 & 2 & 2 & 2 & 0\\
			\hline
			\multicolumn{7}{c}{\textbf{number of cycles}}\\
			\hline
			$\circlearrowright \circ$ & 4 & 4 & 4 & 2 & 2 & 2\\
			\hline
			\multicolumn{7}{c}{\textbf{number of arcs}}\\
			\hline
			$\circ \rightarrow$ & 2 & 2 & 2 & 1 & 1 & 1\\
			\hline
			$\rightarrow \circ$ & 2 & 2 & 2 & 1 & 1 & 1\\
			\hline
		\end{tabular}
		
		\caption{Digraph $\mydirectedGraphsym{D}6_{IV}$}   \label{fig:example2:formiz_modes:digraph_var6_4}
	\end{subfigure}
	
	\caption{Visualizations of the digraphs of Case 2 of Example~\ref{sec:example:mechProp_mech:formizmodes} and the information for their formizations.}   \label{fig:example03:formiz2}
\end{figure}

\paragraph{Case 2} Assume the digraphs illustrated in Figure~\ref{fig:example03:formiz2}.
It is seen from the formization tables of the digraphs that
\begin{itemize}
	\item   All digraphs have the same number of vertices adjacent from and to as well as the same number of cycles over their vertices.
	
	\item   No two digraphs have the same formization in either of the formization modes.
\end{itemize}
From these statement, one concludes that even the two mentioned property measures (i.e.\ the number of vertices adjacent from and to and the number of cycles over vertices of a digraph) cannot (totally) formize a digraph in general.

\subsection{Mechanistic analysis of cognition mechanisms}   \label{sec:example4:mech_analysis_cogmech}
This example is designated to analyze cognition mechanisms through mechanistic characterizations and formizations similar to the previous example as well as to analyze some cases of cognition mechanisms' mechanistic evolution. In addition, this example assumes the digraph $\mydirectedGraphsym{D}$ of example~\ref{sec:example1} and considers that $\mydirectedGraphsym{D}$, $\myfuncground{\mydirectedGraphsym{D}}$, $\myfuncnonground{\mydirectedGraphsym{D}}$, and $\myfuncreciprocalground{\mydirectedGraphsym{D}}$ are fixed entities.

\begin{figure}
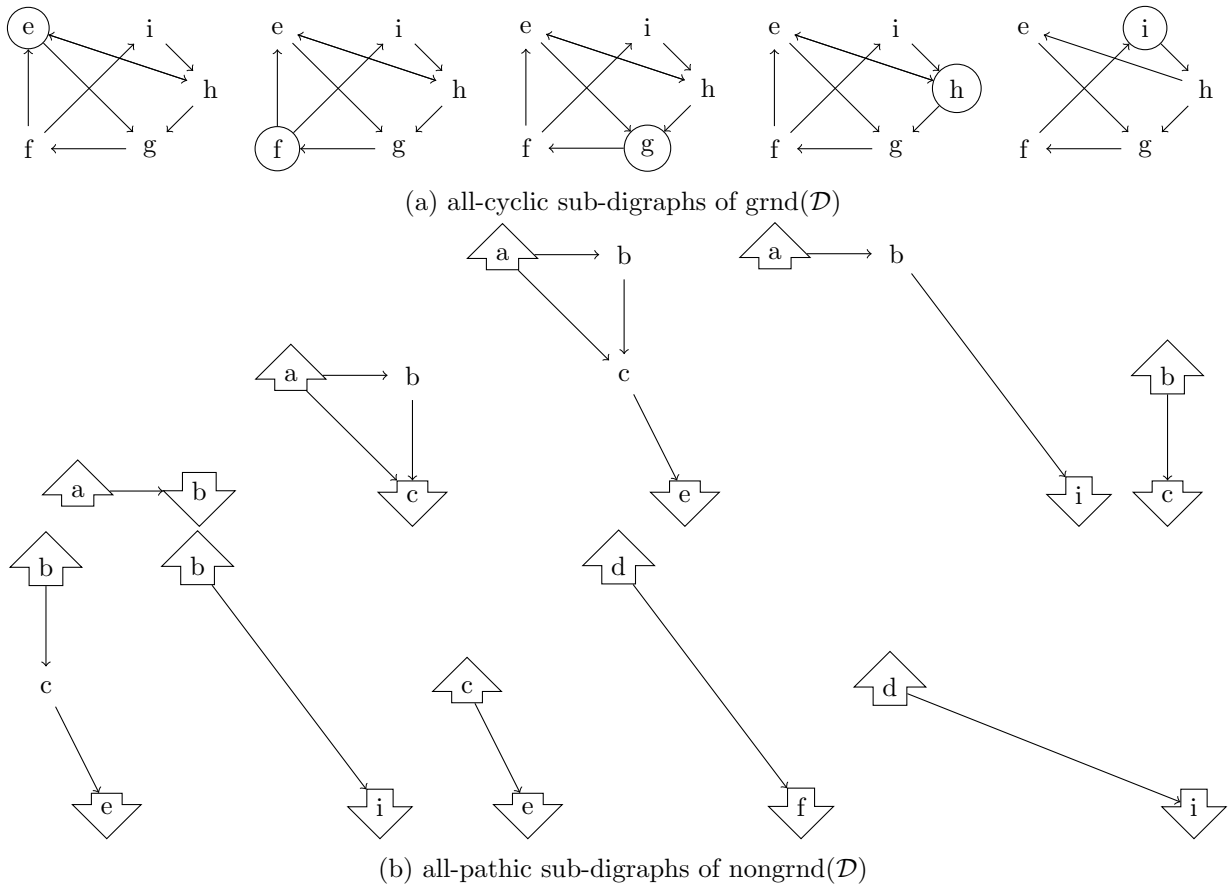

	\centering

	\begin{subfigure}[t]{1.0\textwidth}
		\centering
		
		\input{exmpl4_digraph_unit_e.tikz}
		\input{exmpl4_digraph_unit_f.tikz}
		\input{exmpl4_digraph_unit_g.tikz}
		\input{exmpl4_digraph_unit_h.tikz}
		\input{exmpl4_digraph_unit_i.tikz}
		
		\caption{all-cyclic sub-digraphs of $\myfuncground{\mydirectedGraphsym{D}}$}   \label{fig:example4:cogmech_charz:allcyclicsubdigraphs}
	\end{subfigure}

	\begin{subfigure}[t]{1.0\textwidth}
		\centering
		
		\input{exmpl4_digraph_uniter_ab.tikz}
		\input{exmpl4_digraph_uniter_ac.tikz}
		\input{exmpl4_digraph_uniter_ae.tikz}
		\input{exmpl4_digraph_uniter_ai.tikz}
		\input{exmpl4_digraph_uniter_bc.tikz}
		\input{exmpl4_digraph_uniter_be.tikz}
		\input{exmpl4_digraph_uniter_bi.tikz}
		\input{exmpl4_digraph_uniter_ce.tikz}
		\input{exmpl4_digraph_uniter_df.tikz}
		\input{exmpl4_digraph_uniter_di.tikz}
		
		\caption{all-pathic sub-digraphs of $\myfuncnonground{\mydirectedGraphsym{D}}$}   \label{fig:example4:cogmech_charz:allpathicsubdigraphs}
	\end{subfigure}
	
	\caption{Visualizations of the all-cyclic and the all-pathic sub-digraphs of the digraph $\mydirectedGraphsym{D}$ in Example~\ref{sec:example4:mech_analysis_cogmech} by adopting the standard mechanistic characterization for cognition mechanisms.}   \label{fig:example04:cogmech_characterization}
\end{figure}

For mechanistic characterization of a cognition mechanism, I propose a standard by which a cognition mechanism's self is characterized symbolistically and that cognition mechanism's non-self is characterized connectionistically. Figure~\ref{fig:example04:cogmech_characterization} shows the standard mechanistic characterization of $\mydirectedGraphsym{D}$.

In contrast to mechanistic characterization, it is obviously not possible for mechanistic formization of a cognition mechanism to separate ground and non-ground of a digraph a priori. This is because the digraph is not uniquely specified before the formization process and, consequently, it is not possible to check whether there exists a ground and, if yes, to specify the ground and the non-ground. Formization information of the digraph $\mydirectedGraphsym{D}$ is depicted in Table~\ref{tab:exmple4:cogmech_formiz}.
\begin{table}
	\centering
	
	\caption{The formization information of the digraph $\mydirectedGraphsym{D}$ in Example~\ref{sec:example4:mech_analysis_cogmech}.} \label{tab:exmple4:cogmech_formiz}
	
	\begin{tabular}{| c || c | c | c | c | c | c | c | c | c |}
		\multicolumn{10}{c}{\textbf{number of paths}}\\
		\hline
		$\mypathsymne$ & o1 & o2 & o3 & o4 & o5 & o6 & o7 & o8 & o9\\
		\hline
		\hline
		o1 & 0 & 1 & 2 & 0 & 4 & 6 & 6 & 5 & 5\\
		\hline
		o2 & 0 & 0 & 1 & 0 & 3 & 4 & 4 & 3 & 3\\
		\hline
		o3 & 0 & 0 & 0 & 0 & 1 & 2 & 2 & 2 & 2\\
		\hline
		o4 & 0 & 0 & 0 & 0 & 4 & 3 & 6 & 3 & 2\\
		\hline
		o5 & 0 & 0 & 0 & 0 & 0 & 2 & 2 & 2 & 2\\
		\hline
		o6 & 0 & 0 & 0 & 0 & 2 & 0 & 4 & 2 & 1\\
		\hline
		o7 & 0 & 0 & 0 & 0 & 2 & 1 & 0 & 2 & 1\\
		\hline
		o8 & 0 & 0 & 0 & 0 & 2 & 2 & 2 & 0 & 2\\
		\hline
		o9 & 0 & 0 & 0 & 0 & 2 & 2 & 2 & 1 & 0\\
		\hline
		\multicolumn{10}{c}{\textbf{number of cycles}}\\
		\hline
		over ($\circlearrowright \circ$) & 0 & 0 & 0 & 0 & 4 & 4 & 4 & 4 & 2\\
		\hline
		\multicolumn{10}{c}{\textbf{number of arcs}}\\
		\hline
		from ($\circ \rightarrow$) & 2 & 2 & 1 & 2 & 2 & 2 & 1 & 2 & 1\\
		\hline
		to ($\rightarrow \circ$)   & 0 & 1 & 2 & 0 & 3 & 2 & 2 & 2 & 3\\
		\hline
	\end{tabular}
\end{table}

In the following, I discuss about some cases of evolution of a cognition mechanism in the form of mechanistic modifications of it.
\paragraph{Case 1}
Construct a first modified digraph $\mymodifgraph{\mydirectedGraphsym{D}}1$ (illustrated in Figure~\ref{fig:example02:1:firstmodif}) by assuming a digraph initially identical to $\mydirectedGraphsym{D}$ and then removing the arc $(i, h)$ from it. In $\mymodifgraph{\mydirectedGraphsym{D}}1$, $\myfuncground{\mydirectedGraphsym{D}}$ will satisfy basalness (i.e.\ underlyingness and primitiveness together) but not unifiedness\footnote{Notice that $\myfuncground{\mydirectedGraphsym{D}}$ does not have all required characteristics of a ground in $\mymodifgraph{\mydirectedGraphsym{D}}1$ and, thus, is not a ground of $\mymodifgraph{\mydirectedGraphsym{D}}1$.} because vertex $i$ will not be connected to all of the vertices $e$, $f$, $g$, and $h$. Now, construct a second modified digraph $\mymodifgraph{\mydirectedGraphsym{D}}2$ (illustrated in Figure~\ref{fig:example02:1:secondmodif}) by assuming a digraph initially identical to $\mymodifgraph{\mydirectedGraphsym{D}}1$ and then adding an arc which connects the vertex $i$ to a vertex of $\myfuncground{\mydirectedGraphsym{D}}$, for instance the vertex $g$, to it. Consequently, $\myfuncground{\mymodifgraph{\mydirectedGraphsym{D}}2}$ and $\myfuncnonground{\mymodifgraph{\mydirectedGraphsym{D}}2}$ illustrated in Figure~\ref{fig:example02:1:secondmodif} will be the ground and the non-ground of $\mymodifgraph{\mydirectedGraphsym{D}}2$.
\begin{figure}
	\centering
	
	\begin{subfigure}[b]{0.45\textwidth}
		\centering
		
		\input{exmpl3_digraph_noground.tikz}
		
		\caption{The first modified digraph $\mymodifgraph{\mydirectedGraphsym{D}}1$.}   \label{fig:example02:1:firstmodif}
	\end{subfigure}
	\hspace{0.0cm}
	\begin{subfigure}[b]{0.45\textwidth}
		\centering
		
		\input{exmpl3_digraph_ground.tikz}
		
		\caption{The second modified digraph $\mymodifgraph{\mydirectedGraphsym{D}}2$.}    \label{fig:example02:1:secondmodif}
	\end{subfigure}
	
	\caption{Visualizations of the modified digraphs of Case 1 of Example~\ref{sec:example4:mech_analysis_cogmech}.}   \label{fig:example02:1}
\end{figure}

\paragraph{Case 2}
Construct a first modified digraph $\mymodifgraph{\mydirectedGraphsym{D}}1$ (illustrated in Figure~\ref{fig:example02:3:firstmodif}) 
by assuming a digraph initially identical to $\mydirectedGraphsym{D}$ and then adding a new arc $(e, d)$ to it. In $\mymodifgraph{\mydirectedGraphsym{D}}1$, $\myfuncground{\mydirectedGraphsym{D}}$ will satisfy singleness (i.e.\ underlyingness and unifiedness together) but not primitiveness because the vertex $e$ is connected to the vertex $d$. Without any further modifications, $\mymodifgraph{\mydirectedGraphsym{D}}1$ has a ground. Consequently, $\myfuncground{\mymodifgraph{\mydirectedGraphsym{D}}1}$ and $\myfuncnonground{\mymodifgraph{\mydirectedGraphsym{D}}1}$ illustrated in Figure~\ref{fig:example02:3:secondmodif} will be the ground and the non-ground of $\mymodifgraph{\mydirectedGraphsym{D}}1$.
\begin{figure}
	\centering
	
	\begin{subfigure}[b]{0.45\textwidth}
		\centering
		
		\input{exmpl4_digraph_general.tikz}
		
		\caption{The first modified digraph $\mymodifgraph{\mydirectedGraphsym{D}}1$ in general.\newline}   \label{fig:example02:3:firstmodif}
	\end{subfigure}
	\hspace{0.0cm}
	\begin{subfigure}[b]{0.45\textwidth}
		\centering
		
		\input{exmpl4_digraph_ground.tikz}
		
		\caption{The first modified digraph $\mymodifgraph{\mydirectedGraphsym{D}}1$ and its ground and non-ground.}   \label{fig:example02:3:secondmodif}
	\end{subfigure}
	
	\caption{Visualizations of the modified digraph of Case 2 of Example~\ref{sec:example4:mech_analysis_cogmech}.}   \label{fig:example02:3}
\end{figure}

\paragraph{Case 3}
Construct a first modified digraph $\mymodifgraph{\mydirectedGraphsym{D}}1$ (illustrated in Figure~\ref{fig:example02:2:firstmodif}) 
by assuming a digraph initially identical to $\mydirectedGraphsym{D}$ and then adding a new vertex $\alpha$ and a new arc $(d, \alpha)$ to it. In $\mymodifgraph{\mydirectedGraphsym{D}}1$, $\myfuncground{\mydirectedGraphsym{D}}$ will satisfy uniqueness (i.e.\ primitiveness and unifiedness together) but not underlyingness because vertex $\alpha$ will not be connected to any of vertices of $\myfuncground{\mydirectedGraphsym{D}}$. Now, construct a second modified digraph $\mymodifgraph{\mydirectedGraphsym{D}}2$ (illustrated in Figure~\ref{fig:example02:2:secondmodif}) by assuming a digraph initially identical to $\mymodifgraph{\mydirectedGraphsym{D}}1$ and then adding an arc which connects the vertex $\alpha$ to a vertex of $\myfuncground{\mydirectedGraphsym{D}}$ or $\myfuncnonground{\mydirectedGraphsym{D}}$, for instance the vertex $a$, to it. Consequently, $\myfuncground{\mymodifgraph{\mydirectedGraphsym{D}}2}$ and $\myfuncnonground{\mymodifgraph{\mydirectedGraphsym{D}}2}$ illustrated in Figure~\ref{fig:example02:2:secondmodif} will be the ground and the non-ground of $\mymodifgraph{\mydirectedGraphsym{D}}2$.
\begin{figure}
	\centering
	\begin{subfigure}[b]{0.45\textwidth}
		\centering
		
		\input{exmpl5_digraph_noground.tikz}
		
		\caption{The first modified digraph $\mymodifgraph{\mydirectedGraphsym{D}}1$.}   \label{fig:example02:2:firstmodif}
	\end{subfigure}
	\hspace{0.0cm}
	\begin{subfigure}[b]{0.45\textwidth}
		\centering
		
		\input{exmpl5_digraph_ground.tikz}
		
		\caption{The second modified digraph $\mymodifgraph{\mydirectedGraphsym{D}}2$.}   \label{fig:example02:2:secondmodif}
	\end{subfigure}
	
	\caption{Visualizations of the modified digraphs of Case 3 of Example~\ref{sec:example4:mech_analysis_cogmech}.}   \label{fig:example02:2}
\end{figure}

\paragraph{Case 4}
Construct a first modified digraph $\mymodifgraph{\mydirectedGraphsym{D}}1$ (illustrated in Figure~\ref{fig:example02:4:firstmodif}) 
by assuming a digraph initially identical to $\mydirectedGraphsym{D}$ and then adding a new vertex $\alpha$ and a new arc $(h, \alpha)$ to it. In $\mymodifgraph{\mydirectedGraphsym{D}}1$, $\myfuncground{\mydirectedGraphsym{D}}$ will satisfy unifiedness but not underlyingness nor primitiveness because the vertex $\alpha$ will not be connected to any vertices of $\myfuncground{\mydirectedGraphsym{D}}$ and also the vertex $h$ will be connected to the vertex $\alpha$. Now, construct a second modified digraph $\mymodifgraph{\mydirectedGraphsym{D}}2_I$ (illustrated in Figure~\ref{fig:example02:4:secondmodif_I}) by assuming a digraph initially identical to $\mymodifgraph{\mydirectedGraphsym{D}}1$ and then adding an arc which connects the vertex $\alpha$ to a vertex of $\myfuncground{\mydirectedGraphsym{D}}$, for instance the vertex $i$, to it. Consequently, $\myfuncground{\mymodifgraph{\mydirectedGraphsym{D}}2_I}$ and $\myfuncnonground{\mymodifgraph{\mydirectedGraphsym{D}}2_I}$ illustrated in Figure~\ref{fig:example02:4:secondmodif_I} will be the ground and the non-ground of $\mymodifgraph{\mydirectedGraphsym{D}}2_I$. As another version of a second modification, construct a second modified digraph $\mymodifgraph{\mydirectedGraphsym{D}}2_{II}$ (illustrated in Figure~\ref{fig:example02:4:secondmodif_II}) by assuming a digraph initially identical to $\mymodifgraph{\mydirectedGraphsym{D}}1$ and then adding an arc which connects the vertex $\alpha$ to a vertex of $\myfuncnonground{\mydirectedGraphsym{D}}$, for instance the vertex $a$, to it. Consequently, $\myfuncground{\mymodifgraph{\mydirectedGraphsym{D}}2_{II}}$ and $\myfuncnonground{\mymodifgraph{\mydirectedGraphsym{D}}2_{II}}$ illustrated in Figure~\ref{fig:example02:4:secondmodif_II} will be the ground and the non-ground of $\mymodifgraph{\mydirectedGraphsym{D}}2_{II}$.
\begin{figure}
	\centering
	
	\begin{subfigure}[b]{0.45\textwidth}
		\centering
		
		\input{exmpl6_digraph_noground.tikz}
		
		\caption{The first modified digraph $\mymodifgraph{\mydirectedGraphsym{D}}1$.}   \label{fig:example02:4:firstmodif}
	\end{subfigure}
	\newline
	\begin{subfigure}[b]{0.45\textwidth}
		\centering
		
		\input{exmpl6_digraph_ground_I.tikz}
		
		\caption{The second modified digraph $\mymodifgraph{\mydirectedGraphsym{D}}2_I$.}   \label{fig:example02:4:secondmodif_I}
	\end{subfigure}
	\hspace{0.0cm}
	\begin{subfigure}[b]{0.45\textwidth}
		\centering
		
		\input{exmpl6_digraph_ground_II.tikz}
		
		\caption{The second modified digraph $\mymodifgraph{\mydirectedGraphsym{D}}2_{II}$.}   \label{fig:example02:4:secondmodif_II}
	\end{subfigure}

	\caption{Visualizations of the modified digraphs of Case 4 of Example~\ref{sec:example4:mech_analysis_cogmech}.}   \label{fig:example02:4}
\end{figure}

\section{Discussions and prospects}   \label{sec:discussions_prospects}
Three aspects of this paper: the used terminology, the cognition mechanism, and the framework of defining and analyzing cognition mechanisms are discussed and the prospects of applying and enhancing the framework are expressed in this section.

I have primarily adopted a \emph{definitive} approach to talk about mechanisms. This definitive approach is a symbolistic characterization of mechanisms in its core which is the used terminology. Therefore, terms of the terminology define themselves, which means that the extensional definitions of the terms are essentially in the form of a collection of recursive statements. This collection's statements may be at their most abstract form which cognizing them do not require a priori mutual intensions (= intensional entities) between the intender and the extender of them (which, here, are a reader of this paper and I, respectively). In practice, an intender of extensional definitions may face situations in which it is needed to back-and-forth refer to the definitions in order to identify the recursive statements, and consequently, to clearly cognize them. The proposed (extensional) definitions of the terms used in this paper are abstract and so is the definition of a mechanism. To make it more concrete, the terms `node' and `direct link' might be considered as `matter' and `composive/decomposive direction', `entity' and `causative/effectuative direction', `state' and `relatuative direction', 'position' and `connectative direction', etc.\ which are more connected to reality. In the features of a human self which are brought in the introduction and are related to the properties of a cognition self in Section~\ref{sec:cognition_base}, `matter' and `decomposive direction' is considered for `node' and `direct link', which means that one should translate ``There is a direct link from node $x$ to node $y$'' to ``There is a decomposive direction from matter $x$ to matter $y$''. Similarly, the term `intramission' is considered as `decomposition'. It should be noted that the substituted term is a partial action in general and may not tell all the truth about a situation. In addition, in Section~\ref{sec:cognition_base}, the term `speculation' is considered for `intramission' (which is initiated from a node of the cognition self) when comparing a cognition mechanism's properties with another set of human cognition features. This usage of `speculation' implies consideration of `node' and `direct link' of the cognition self as `presentational entity' and `speculative direction'.

In addition to the approach explained in the previous paragraph, there is a \emph{descriptive} approach. A mechanically similar explanation of this approach is
\begin{quote}
	I have secondarily adopted a descriptive approach to talk about mechanisms. This descriptive approach is a connectionistic characterization of mechanisms in its core which is the used terminology. Therefore, terms of the terminology describe other terms, which means that the extensional descriptions of the terms are essentially in the form of a collection of cursive statements. This collection's statements may be at their most concrete form which cognizing them do require a priori mutual intensions between the intender and the extender of them (which, here, are a reader of this paper and I, respectively). In practice, an intender of extensional descriptions may face situations in which it is needed to forwardly refer to other descriptions in order to realize the cursive statements, and consequently, to clearly cognize them.
\end{quote}
Accordingly, an instance of description of a mechanism would be
\begin{quote}
	There is a collection of boxes with unique colors. Each box has one-way wire to one or one-way wires to several other boxes. Also, there is a head which, at a time, can position itself on a box, can see the color of a box, and can move to another box via one of the one-way wires.
\end{quote}
From this description, a mechanism is cognized as the collection of colored and one-way wired boxes and a head with the stated capabilities.

As defined in the introduction, a cognition mechanism is a mechanism which has a base and which intramits in the base. The base and the intramission in the base are translated as `self' and `speculation' of a human in the paper. Yet, human cognition has three other features: memory, ception/action, and will which the defined (semi-quasi-pseudo-)cognition does not. For example, situatedness of human cognition involves embodiment, embeddedness, enaction, affect, and extendedness which the defined cognition has the first two but not the other ones because they necessitate the three features. For instance, an extended cognition thesis says that a mind extends to its environment (or even other minds) in order to cognize entities, which the defined cognition mechanism cannot be extended as such, again, because it does not possess the three features. Nonetheless, the defined cognition mechanism is not only compatible with the situatedness of human cognition but it can also be consistent with them by enhancing the cognition mechanism to incorporate memory, ception/action, and will. As practical evidences of the potential of such enhancement, a) the possibility of (random) initiation from as well as (random) termination to a node of the self in the intramissions of a cognition mechanism may be defined as a requirement of having (free) will, b) the possibility of joining intramissions may be defined as a requirement of having memory, and c) the possibility of evolution of a cognition mechanism (e.g.\ those exemplified in \ref{sec:example4:mech_analysis_cogmech}) may be defined as a requirement of having ception/action. As soon as cognition mechanisms have all human cognition features, a cognition mechanism can observe a cognition mechanism. As a result, in my opinion, an observing cognition mechanism sees a cognition mechanism as a mechanism having a) the five features of mind: consciousness, intentionality, freedom (of will), teleology, and normativity (see Pernu~\cite{Pernu_2017}) or b) a self characterized through the five axioms of IIT: intrinsic existence, composition, information, integration, and exclusion (see Tononi~\cite{Tononi2015}). A meta-mechanation defined in \ref{sec:metainfra_mechs} is an instance of such an observation for semi-quasi-pseudo-cognition mechanisms. 

The proposed framework is sufficiently abstract and general not only to incorporate all human cognition features but also to introduce \emph{super-cognition} (i.e.\ the cognition processes which humans are able to do but are not natural of humans) and \emph{hyper-cognition} (i.e.\ the cognition processes which humans are not able to do but machines might do). A mechanism whose a) nodes and/or direct links possess more than one real label, direct links possess more than two ends, intramissions possess more than one initial/terminal node, constituents also include \emph{nodilinks} (i.e.\ the entities whose characteristics are positioned between those of nodes and direct links and can be constructed through a meta-mechanation), or a combination of these is an example of a super-cognition mechanism. And, a super-mechanism in which infinite and/or uncountable entities are correspondingly substituted for finite countable entities in its definition is a hyper-cognition mechanism. For example, a hyper-cognition mechanism may possess a  real number\footnote{The term `real' in `mathematical real number' should not be confused with `real' in `real labels of nodes'. The former has a rigorous definition in mathematics and the latter is the opposite of `imaginary' in philosophy.} of nodes and direct links where the direct links relate a real number of nodes to each other.

\section{Summary}
This paper proposes a framework of defining, modeling, and analyzing cognition mechanisms. A `cognition mechanism' should incorporate `cognition' and `mechanism'. Cognition is computation and representation by mind (i.e.\ a system which has a base named self); and, a mechanism has constituents and emergents. Thus, a cognition mechanism is defined as a mechanism having a base among its constituents and a process which engages that base among its emergents. A mechanism is defined using a proposed terminology and is modeled by a mathematical digraph and walks in it. The characteristics of a mechanism are clarified through meta-cognitive justifications. As pointed, cognition mechanisms are a class of mechanisms having self. The conditions of existence of a self in a mechanism are proposed. It is assessed that the resulted cognition mechanism satisfies features of human cognition. Furthermore, meta-, infra-, and iso-mechanisms are introduced, which are utilized in the analysis of mechanisms. Then, examples of analyzing cognition mechanisms in the framework are given for a more concrete understanding of them. Finally, the terminology, the cognition mechanisms, and the framework are discussed and prospects of development of the framework are briefly depicted.

\clearpage
\bibliographystyle{unsrt}
\bibliography{my_biblio}

\clearpage
\appendix

\section{Definition of a mechanism from an epistemic first-person view and mathematical model of the mechanism}   \label{apdx:mech_terms_firstperson}
In this section, the definition of a mechanism is stated by using an epistemic first-person terminology. The terminology is correspondent with the mathematical terminology as in Table~\ref{tab:cog_terminology_firstperson}. In addition to the terminology and for the sake of ease of reading and understanding, a) the definition of the mechanism is rephrased and partially reordered and b) some meta-mathematical terms are introduced.

Constituents of a mechanism are noti and their adjacencies. A \emph{notus} is a labeled point (or a mathematical point). Imagine noti\footnote{the plural of notus} which, in a way, are \emph{adjacent} to each other. Therefore, for each notus, there is a \emph{neighborhood} which is all noti being adjacent to it. A notus associated with its neighborhood is called a \emph{notion} and a collection of notions is called a \emph{notional world}. If noti and their adjacencies are \emph{countable}, they are respectively called a collection of \emph{notes} and \emph{immediate dispositions}. I postulate that a) all noti and their adjacencies are countable and b) there can always be an immediate disposition from a note to each of its neighborhood's notes. With these postulates, the notional world definitionally becomes a \emph{conceptual world}. A \emph{concept} is defined as a fragment of the conceptual world\footnote{Notice that a fragment of a world is either nothing, a piece of the world, or the entire world.}. A \emph{uniation} of concepts is defined as a concept composed of a collection of all concepts' notes together with a collection of all concepts' immediate dispositions. If we make numbered notes according to consecutive immediate dispositions of the conceptual world, the collection of the numbered notes is an \emph{instance}. A collection of instances is called an \emph{instantual world}. In an instance, the note with the smallest and the largest number are called the \emph{initial} and the \emph{terminal} notes, respectively. If the initial and the terminal notes of an instance are the same, the instance is called an \emph{object} (over the initial or the terminal note); otherwise, it is a \emph{relation} (from the initial to the terminal note). If the notes of an object except the terminal note or the notes of a relation are distinct from each other, that object or relation is \emph{primary}. If the collection of numbered notes of an instance is converted to a collection of notes without numbers and a collection of immediate dispositions according to the consecutive numbers, one obtains the \emph{underlying concept} (of the instance). A conceptual world is \emph{objectival over a note} if at least an object over that note can be instantiated from a concept of it; and, a conceptual world is \emph{relatival between two notes} if at least a relation between the two notes can be instantiated from a concept of it. Similarly, a conceptual world is \emph{strongly} (primary-)objectival or \emph{strongly} (primary-)relatival if it is (primary-)objectival over or (primary-)relatival between its every note, respectively. A strongly relatival conceptual world is strongly primary-objectival but a strongly primary-objectival conceptual world is not necessarily strongly relatival. In a conceptual world, a) a \emph{symbol} over note $x$ (or shortly a symbol $x$) is a concept which is the uniation of underlying concepts of all possible primary objects over note $x$ of the conceptual world and b) a \emph{connection} from note $x$ to note $y$ (or shortly a connection $x$--$y$) is a concept which is the uniation of underlying concepts of all possible primary relations from note $x$ to note $y$ of the conceptual world. An instance can be combined to another instance and make a new instance only if the terminal note of the first instance is the same as the initial note of the second instance. The procedure of \emph{combination} of instances is done by a) changing the second instance's initial note's number so that it will be equal to the first instance's terminal note's number, b) changing the rest of the second instance's notes' numbers according to their sequence, and c) collecting the first instance's and the modified second instance's numbered notes together.
\begin{table}
	\centering
	\caption{The used epistemic first-person terminology in definition of a mechanism corresponding to the mathematical terminology}   \label{tab:cog_terminology_firstperson}
	\begin{tabular}{p{6cm} | p{6cm}}
		\hline
		Cognitive term & Mathematical term\\
		\hline
		notus & point\\
		adjacency & (mathematical) relation in the form of pairs of objects\\
		neighborhood & neighborhood topology pair\\
		collection of neighborhoods & neighborhood topology\\
		notion & neighborhood topological spatium\\
		notional world & neighborhood topological space\\
		note & vertex\\
		immediate disposition & arc\\
		conceptual world & digraph\\
		concept & sub-digraph\\
		instance & walk\\
		object & closed walk\\
		relation & open walk\\
		simple object & cycle\\
		simple relation & path\\
		underlying concept of an instance & traversed sub-digraph of a walk\\
		uniation & union\\
		symbol & all-cyclic sub-digraph\\
		connection & all-pathic sub-digraph\\
		(strongly) primary-objectival & (strongly) cyclic\\
		(strongly) relatival & (strongly) connected
	\end{tabular}
\end{table}

The process of a) finding the underlying concept of an instance is called \emph{conception} (of the instance) and b) making an instance underlay by a concept is called \emph{instantiation} (of the concept). Instantiation itself is divided into 1) choosing an initial note, called \emph{initiation}, 2) sequentially following immediate dispositions of notes, called \emph{sequention}, and 3) choosing a terminal note, called \emph{termination}. If one observes an instance, does a conception of that instance, considers the underlying concept as a conceptual world, and defines an instantual world on that conceptual world, he/she has performed a \emph{ceptive mechanation}. On the other hand, if one imagines a concept, does an instantiation of that concept, considers the obtained instance as an instantual world, and defines a conceptual world from that instantual world, he/she has performed an \emph{active mechanation}. If a mechanation (either ceptive or active) gives something that is trivially equivalent to what it has started from, it is \emph{steady}. For example, a ceptive mechanism is steady if it gives an instantual world which is equivalent to the instance which is observed at the start of mechanation; or an active mechanism is steady if it gives a conceptual world which is equivalent to the concept which is imagined at the start of mechanation. The concept, the instance, and the way they are transformed into each other in a (steady) mechanation are altogether called a \emph{(stable) mechanism}.

\clearpage
\section*{CRediT (Contributor Roles Taxonomy)}
\textbf{Amir Fayezioghani} Conceptualization; Data curation; Formal Analysis; Funding acquisition; Investigation; Methodology; Project administration; Software; Validation; Visualization; Writing -- original draft; Writing -- review \& editing.\footnote{Only the major roles applicable to the author are listed although all other CRediT roles (refer to \url{https://credit.niso.org/}) are also applicable in a way.}

\end{document}